\documentclass[runningheads]{llncs}

\usepackage{eccv}

\usepackage{eccvabbrv}
\usepackage{graphicx}
\usepackage{booktabs}
\usepackage{subcaption}
\usepackage{xr-hyper}

\usepackage{fvextra}
\usepackage{array}
\usepackage{multirow}

\usepackage[accsupp]{axessibility}  

\usepackage{hyperref}

\usepackage{orcidlink}

\usepackage{marvosym} 

\newcommand{\symbolfootnotetext}[2]{%
  \begingroup
  \renewcommand{\thefootnote}{#1}%
  \footnotetext{#2}%
  \endgroup
}

\begin{document}
\raggedbottom
\title{H-Adapter: Pose-Robust Hairstyle Transfer via Attention-Derived, Source-Aligned Hair Masks}

\titlerunning{H-Adapter for Pose-Robust Hairstyle Transfer}


\author{Seulgi Jeong\orcidlink{0009-0009-7588-7981}\textsuperscript{*} \and
Yunseong Cho\orcidlink{0009-0007-2131-9909} \and
Sanghun Park\orcidlink{0000-0002-7364-2915}\textsuperscript{\Letter} }

\authorrunning{S.~Jeong et al.}

\institute{SNOW Corp., 
Republic of Korea\\
\email{wjdtmfrl3682@gmail.com, \{yunseong.cho,sanghun.park\}@snowcorp.com}
}


\maketitle
\symbolfootnotetext{*}{Work done during an internship at SNOW Corp.}
\symbolfootnotetext{\Letter}{Corresponding author.}

\begin{abstract}
Hairstyle transfer has practical applications such as virtual try-on, yet remains challenging when the source and reference exhibit large head-pose discrepancies. We propose H-Adapter, which improves pose robustness by training with a region-specific loss that disentangles hair and non-hair objectives and thereby induces spatially disentangled cross-attention, from which a source-aligned hair edit mask is derived to guide diffusion-based inpainting. Experiments on pose-agnostic and pose-different subsets demonstrate strong quantitative results, including the best FID, $\mathrm{FID}_{\mathrm{CLIP}}$, and CLIP-I under pose differences, while maintaining competitive non-hair preservation and improving qualitative fidelity to fine-grained reference hairstyle details. Beyond source-conditioned transfer, H-Adapter supports practical extensions including reference-guided text-to-image generation, auxiliary prompt-based hair color control, and compatibility with an identity-preserving IP-Adapter variant. We also introduce a VLM-as-a-judge protocol and observe consistent gains in hairstyle faithfulness, non-hair preservation, and artifact quality.
\end{abstract}

\begin{figure}[t]
    \centering
    \includegraphics[width=\linewidth]{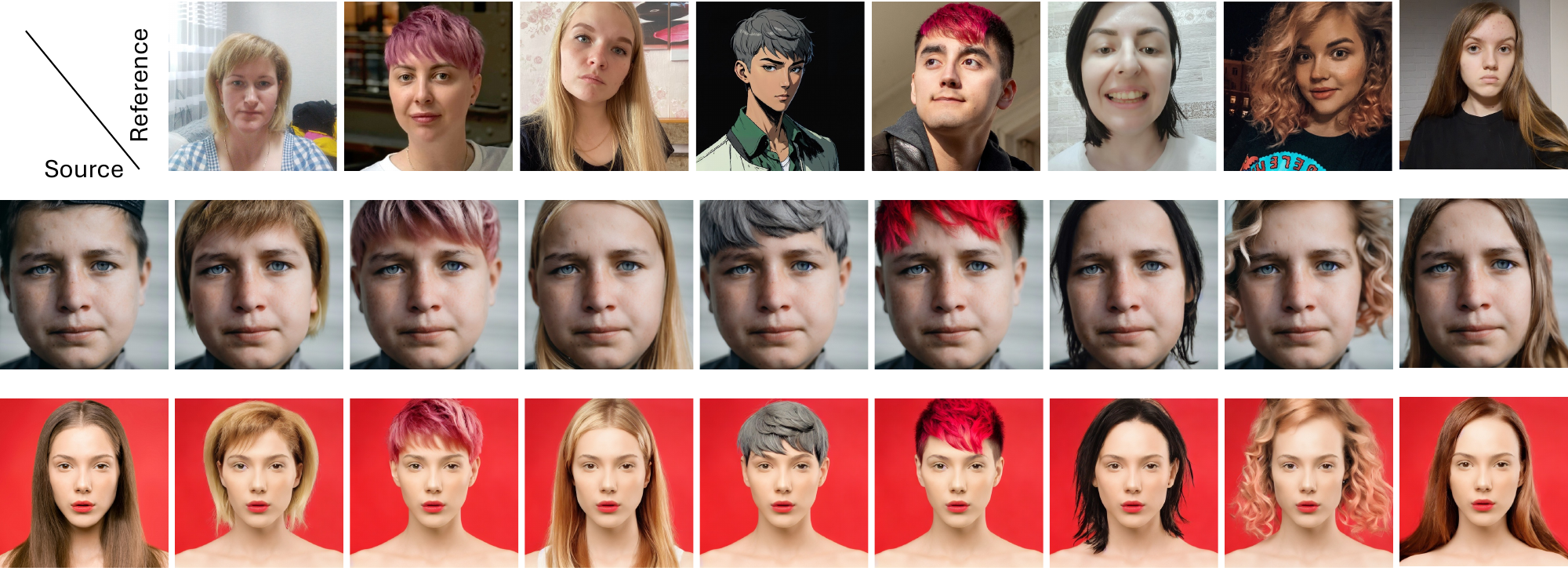}
    \caption{Qualitative results of H-Adapter (Ours) on EasyPortrait~\cite{kvanchiani2023easyportrait} and web-crawled images. H-Adapter preserves image quality and reference hairstyle features under diverse, unconstrained conditions. Source and license information for the Pexels, Pixabay, and Unsplash images used in this figure is provided in the supplementary material.}
    \label{fig:in-the-wild}
\end{figure}

\section{Introduction}
\label{sec:intro}

Recent progress in image synthesis and editing has led to higher-fidelity generation and more precise local image manipulations.
Hairstyle transfer has emerged as an important image editing task with practical applications such as virtual hairstyle try-on.
Given a source and a reference image, the goal is to transfer the reference hair color and shape while preserving non-hair content (\eg, identity, background, and clothing).
This task imposes region-dependent objectives: the model must reflect reference hairstyle features in hair regions while preserving the source content in non-hair regions. This coupling of conflicting objectives makes it difficult to localize edits without degrading non-hair content.

A key challenge arises from the non-rigid, pose-dependent nature of hair: changes in head pose can substantially shift its spatial placement and silhouette, so successful transfer requires determining where hair should appear on the source head geometry and in what approximate shape, especially under large source--reference pose discrepancies.

GAN-based approaches~\cite{saha2021loho, tan2020michigan, zhu2021barbershop, wei2022hairclip, wei2023hairclipv2, nikolaev2024hairfastgan, kim2022style, chung2022hairfit} have driven much of the recent progress in hairstyle transfer.
These methods typically rely on aligned face datasets, which can be effective in controlled settings but limit robustness in unconstrained imagery.
To improve generalization under unconstrained conditions, diffusion-based approaches~\cite{zeng2024hairdiffusion, zhang2025stable, chung2025preserve} have also been explored.
However, under pose mismatch, two failure modes remain common: (1) inaccurate hair-region localization, leading to misplaced or over-extended edits; and (2) limited reproduction of fine-grained reference structure and texture, even when global placement is reasonable.

Existing methods often fail to localize the editable hair region under large pose mismatch. We therefore seek \emph{source-aligned spatial guidance} that specifies where hair should be synthesized on the source head geometry, and use it to constrain diffusion-based inpainting~\cite{rombach2022high} to the hair region.

We propose H-Adapter, a hair-aware adapter trained with a novel \emph{region-specific loss} to spatially disentangle hair and non-hair objectives, built upon IP-Adapter~\cite{ye2023ip}. Concretely, the loss encourages the model to learn reference hairstyle appearance only within hair regions, while suppressing reference conditioning in non-hair regions by regularizing them toward the pretrained diffusion model’s predictions.
This training yields cross-attention maps with increased spatial separation between hair and non-hair content, from which we derive a coarse, source-aligned hair mask that guides diffusion-based inpainting.
\Cref{fig:in-the-wild} provides representative in-the-wild examples, illustrating the practical applicability of H-Adapter under source--reference pose discrepancies and other unconstrained image variations.

Beyond improving localization under pose mismatch, the adapter-based design also enables flexible inference-time use: H-Adapter supports auxiliary prompt-based hair-color control in hairstyle transfer, general prompt-driven generation, and compatibility with identity-preserving IP-Adapter variants. Our contributions are summarized as follows:
\begin{itemize}
    \item We introduce a region-specific loss that disentangles hair and non-hair objectives for faithful reference hairstyle transfer.
    \item We derive a source-aligned coarse attention mask from H-Adapter cross-attention to guide diffusion inpainting with pose and shape consistency.
    \item We demonstrate plug-and-play inference, including source-free reference-guided text-to-image generation, optional text-guided hair-color control, and compatibility with other IP-Adapter variants.
\end{itemize}

We achieve strong results under pose mismatch across quantitative, qualitative, VLM-as-a-judge, and human preference evaluations.

\begin{figure}[tb]
\centering

\begin{subfigure}[t]{0.35\textwidth}
  \centering
  \includegraphics[width=\linewidth]{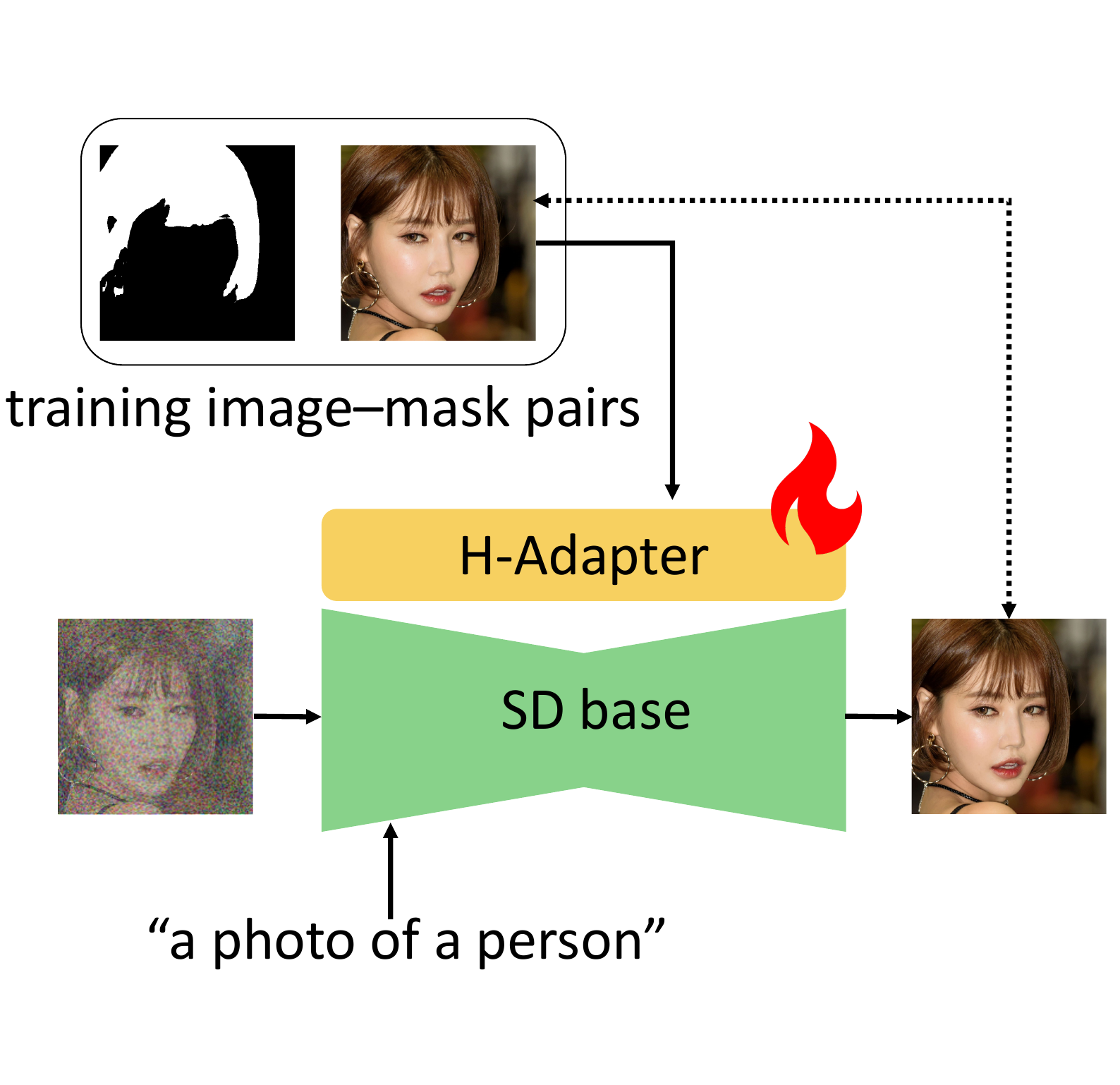}
  \caption{Training pipeline of H-Adapter. The adapter is optimized with a region-specific objective using training image-mask pairs to encourage hair-region-focused conditioning.}
  \label{fig:method-overview-a}
\end{subfigure}\hfill
\begin{subfigure}[t]{0.60\textwidth}
  \centering
  \includegraphics[width=\linewidth]{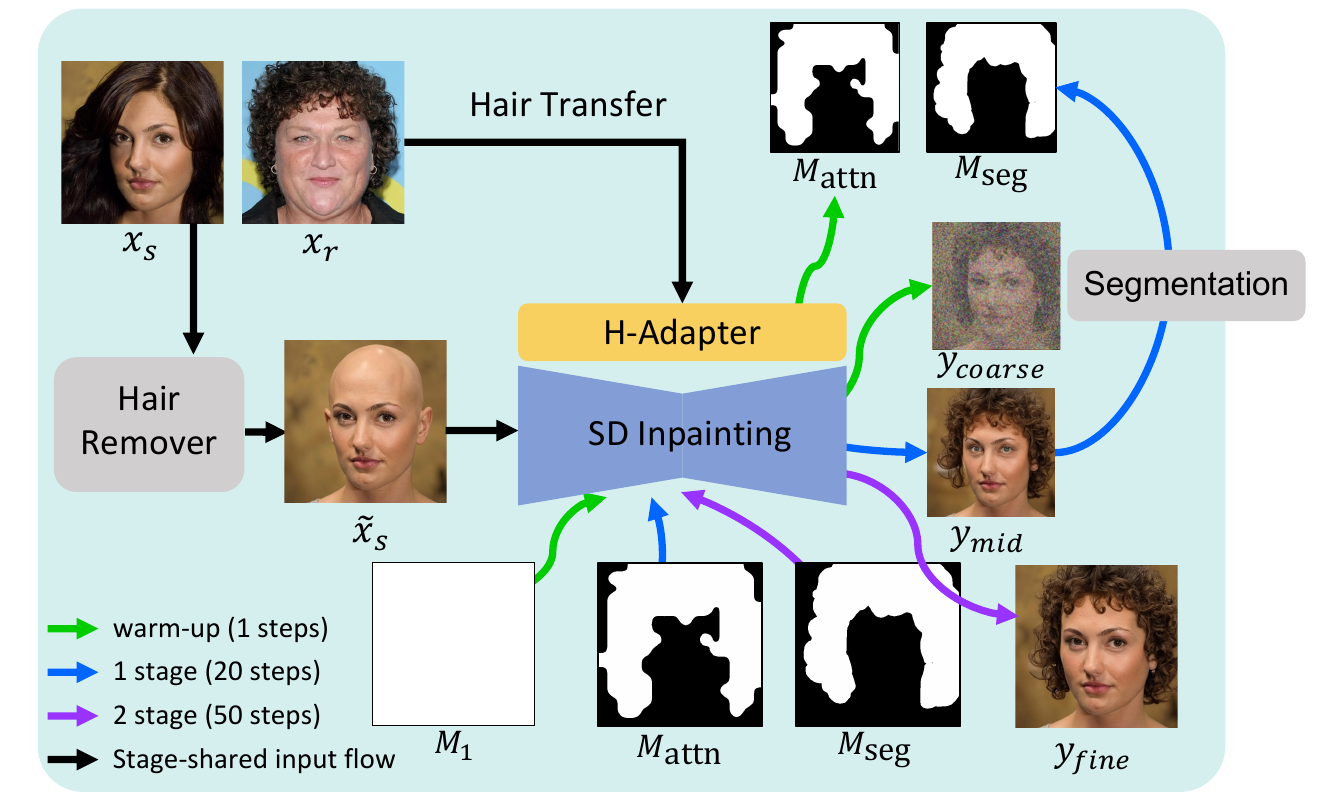}
  \caption{Two-stage inference pipeline. HairRemover~\cite{bfl2026flux2klein} converts the source image $x_s$ into a hair-removed base image $\tilde{x}_s$. Green arrows denote coarse mask extraction ($M_{\mathrm{attn}}$), blue arrows denote intermediate generation ($y_{\mathrm{mid}}$) and segmentation ($M_{\mathrm{seg}}$), and purple arrows denote final inpainting to produce $y_{\mathrm{fine}}$.}

  \label{fig:method-overview-b}
\end{subfigure}

\caption{Overview of our method. We train an H-Adapter with a region-specific objective and apply a source-aligned coarse attention mask to localize diffusion inpainting for reference-guided hairstyle transfer.}
\label{fig:method-overview}
\end{figure}

\section{Related Work}
\label{sec:related-work}

\subsection{Hairstyle Transfer}
\label{sec:related-work:related-hair-transfer}

Prior work on hairstyle transfer has largely focused on structured control signals to improve editability and fidelity.
Several methods explicitly disentangle hair attributes to enable more controllable editing~\cite{tan2020michigan, saha2021loho, zhu2021barbershop, nikolaev2024hairfastgan, wei2023hairclipv2}. Text-driven control is further explored in HairCLIP~\cite{wei2022hairclip}.

More recent approaches expand the conditioning space for practical use, ranging from diffusion-based text conditioning for multi-color hair editing~\cite{zeng2024hairdiffusion} to HairCLIPv2~\cite{wei2023hairclipv2}, which supports multi-modal, user-interactive controls via proxy feature blending; a follow-up work~\cite{wei2026unifying} generalizes this proxy-based editing paradigm, including 3D-aware hair editing.

However, large head-pose differences between the source and reference images remain challenging.
To address this, prior works propose various alignment and pose-conditioning strategies.
Barbershop~\cite{zhu2021barbershop} performs GAN-based image compositing with segmentation masks.
Style-Your-Hair~\cite{kim2022style} aligns hairstyles by optimizing a target-hair latent code toward the source pose.
HairFiT~\cite{chung2022hairfit} leverages multi-view datasets and flow-based alignment to better handle viewpoint changes, while HairNet~\cite{zhu2022hairnet} further addresses hairstyle transfer with pose changes by removing the source hair and transferring a pose-aligned reference hairstyle.
HairFastGAN~\cite{nikolaev2024hairfastgan}, a later GAN-based method designed to mitigate head-pose mismatch, introduces a Rotate Encoder that transforms the face latent to enable pose-aware transfer.
Stable-Hair~\cite{zhang2025stable} exploits video data by transferring hairstyles across frames of the same identity to increase pose robustness.
HairFusion~\cite{chung2025preserve} injects pose cues (\eg, DensePose~\cite{guler2018densepose}) into attention to encourage pose-aligned hair generation.

Despite these efforts, handling large pose gaps between the source and reference images remains challenging.
Several existing approaches mitigate pose variation through pose-aware alignment, latent transformations, or auxiliary pose cues~\cite{chung2022hairfit,kim2022style,nikolaev2024hairfastgan,chung2025preserve}. In contrast, our method derives a source-aligned coarse edit region from cross-attention maps induced by our region-specific loss and uses it to guide diffusion inpainting for pose-consistent hairstyle transfer.

\subsection{Image Editing}
\label{sec:related-work:related-editing}

\subsubsection{Controlling Attention for Editing}

Recent advances in text-to-image diffusion models~\cite{rombach2022high, saharia2022photorealistic, podell2023sdxl, nichol2021glide, ramesh2022hierarchical} have improved generation fidelity and enabled more precise local control in image editing.
In particular, several works manipulate attention to localize edits and preserve structure.
Prompt-to-Prompt~\cite{hertz2022prompt} performs text-driven editing by manipulating cross-attention maps to localize changes across denoising steps, without relying on explicit masks.
Plug-and-Play methods~\cite{tumanyan2023plug} inject self-attention from a guidance image into the generation process to preserve structure in image-to-image translation.

\subsubsection{Mask Generation from Attention}

Several works derive spatial masks from attention maps to localize edits.
MasaCtrl~\cite{cao2023masactrl} separates foreground and background using cross-attention and constrains each region to attend only to the corresponding source region.
DiffEdit~\cite{couairon2022diffedit} derives edit masks from differences in diffusion noise predictions, and InstDiffEdit~\cite{zou2024towards} selects representative tokens and aggregates token relevance to generate masks in real time.

Our approach also leverages attention-derived masks, but targets hair transfer under pose mismatch.
Unlike prior work on general edit localization, we derive a source-aligned hair mask from cross-attention maps that separate hair and non-hair content.


\section{Method}

\subsection{Overview}
\label{sec:method:method-overview}
An overview of our framework is shown in \cref{fig:method-overview}.
Given a source image $x_s$ and a reference image $x_r$, our goal is to transfer the reference hairstyle to the source while preserving non-hair regions.
Our method consists of three parts: 
(i) training IP-Adapter as H-Adapter with a region-specific objective that decouples hair-region reference learning from non-hair consistency regularization (\cref{sec:method:region-specific-loss}),
(ii) an inference pipeline that leverages a source-aligned coarse attention mask to restrict diffusion inpainting~\cite{rombach2022high} to the target hair region (\cref{sec:method:inference}), and (iii) extensions enabled by the plug-and-play compatibility of H-Adapter with text prompts and other adapters (\cref{sec:method:extensions}).

\subsection{Training the H-Adapter}
\label{sec:method:region-specific-loss}

We adopt a pretrained IP-Adapter~\cite{ye2023ip} to inject reference hairstyle features through cross-attention during diffusion denoising.
However, directly applying a standard IP-Adapter to hairstyle transfer is non-trivial, as it is trained to inject conditioning features globally and thus tends to affect non-hair regions as well.
As shown in Fig.~\ref{fig:exp:step_wise}(g) (rows 2--3), directly applying IP-Adapter often fails to transfer hairstyles reliably, leading to incomplete or incorrect hair edits.
To confine the conditioning effect to hair, we train the adapter with a region-specific objective, as illustrated in \cref{fig:method-overview-a}.

We train the adapter using triplets $(x_i,x_j, M_i)$, where $x_i$ is the target image used for diffusion denoising, $x_j$ is the reference image used as the H-Adapter condition, and $M_i$ is the binary hair mask of $x_i$.
Since paired images of the same identity with controlled hairstyle changes are difficult to obtain in practice, we train with a mixture of self-supervised and video-based supervision. Specifically, for a subset of samples, we adopt a self-supervised setup where the same image serves as both the reference and the target (\ie, $x_j=x_i$) providing direct reconstruction supervision. 
For the remaining samples, we construct reference--target pairs from videos such that $x_j$ and $x_i$ depict the same identity but correspond to different frames (\ie, $x_j\neq x_i$), exposing the adapter to natural hairstyle variations while preserving identity consistency.

\subsubsection{Region-Specific Training Objective}
The region-specific objective is designed to impose region-dependent behavior on the H-Adapter across hair and non-hair regions.
In hair regions, the masked denoising loss encourages the adapter to incorporate reference hairstyle cues into the denoising prediction.
In non-hair regions, its reference-conditioning effect should be suppressed so that the prediction remains consistent with the pretrained diffusion model.
By separating these two objectives, the loss encourages the reference branch to become hair-focused and to produce spatially disentangled attention maps between hair and non-hair content, which can later be used as a source-aligned localization cue during inference.
We formalize this design with two complementary loss terms.

First, we apply the standard diffusion denoising loss only within the hair region:
\begin{equation}
L_{\mathrm{hair}}=
\mathbb{E}_{z,\epsilon\sim\mathcal{N}(0,I),t}
\left[\left\|(\epsilon-\epsilon_{\theta_{\mathrm{H}}}(z_t,t,c_t,c_i))\cdot M_i\right\|_2^2\right],
\end{equation}
where $z_t$ is the noisy latent at timestep $t$, $\epsilon\sim\mathcal{N}(0,I)$ is Gaussian noise, $c_t$ and $c_i$ denote the text and reference-image conditions, and $\epsilon_{\theta_{\mathrm{H}}}$ is the noise predictor augmented with the H-Adapter.

Second, to suppress the H-Adapter's influence on non-hair regions, we regularize the model to match the pretrained diffusion model's noise prediction outside the hair mask:
\begin{equation}
L_{\mathrm{non\text{-}hair}}=
\mathbb{E}_{z,\epsilon\sim\mathcal{N}(0,I),t}
\left[\left\|(\epsilon_{\theta_{\mathrm{H}}}(z_t,t,c_t,c_i)-\epsilon_{\theta_0}(z_t,t,c_t))\cdot (1-M_i)\right\|_2^2\right],
\end{equation}
where $\epsilon_{\theta_0}$ is the baseline noise predictor without the H-Adapter.

The final training objective is
\begin{equation}
L = L_{\mathrm{hair}} + \lambda_{\mathrm{non\text{-}hair}}\,L_{\mathrm{non\text{-}hair}},
\end{equation}
where $\lambda_{\mathrm{non\text{-}hair}}$ controls the strength of the non-hair consistency term and is set to $0.1$ in all experiments.

\subsection{Inference-Time Hairstyle Transfer}
\label{sec:method:inference}
Since the proposed training objective is independent of the inpainting process, we train the H-Adapter on the base text-to-image backbone. The learned adapter is then directly transferred to the inpainting backbone used by our coarse-to-fine inference pipeline. We obtain the hair-removed base image $\tilde{x}_s$ from the source image $x_s$ by prompting a pretrained instruction-based image editing model~\cite{bfl2026flux2klein}.
We then perform a two-stage coarse-to-fine inpainting process, as shown in \cref{fig:method-overview-b}, to obtain the final hairstyle transfer result. 

\subsubsection{Source-Aligned Coarse Attention Mask}
A key observation is that the proposed region-specific objective yields cross-attention maps with spatial separation between hair and non-hair regions.
\Cref{fig:exp:step_wise}(b) visualizes the cross-attention maps for all $K$ tokens $(t_0,\ldots,t_{K-1})$.
We use the token that consistently attends to non-hair regions as a separator token $t_s$, and aggregate the remaining token maps to obtain an attention-derived coarse inpainting mask:
\begin{equation}M_{\mathrm{attn}}=\mathrm{Binarize}\!\left(\sum_{k\neq s} CA(t_k)\right),
\end{equation}
where $CA(t_k)$ denotes the cross-attention map of token $t_k$.
In our implementation, $K=16$ and we set $s=8$ based on the validation protocol described in Appendix~\ref{sup:attn} of the Supplementary Material.
The resulting attention-derived mask is used in two ways.
First, it serves as the inpainting mask $M_\mathrm{attn}$ in the two-stage inference pipeline.
Second, the same attention-based localization is recomputed at each denoising timestep for reference gating, as described next.

\subsubsection{Per-timestep Reference Gating}
During inference, we recompute an attention-derived spatial mask $M_t$ using the same separator-token aggregation described above and use it to gate the H-Adapter reference branch in cross-attention.
Following an existing masked reference-injection cross-attention implementation, \footnote{\url{https://github.com/huggingface/diffusers}}
we compute
\begin{equation}
    \label{eq:method:ref-gated-attn}
    Z = \mathrm{softmax}\!\left(\frac{QK^{\top}}{\sqrt{d}}\right)V
    + \lambda \Bigl(M_t \odot \mathrm{softmax}\!\left(\frac{Q{K'}^{\top}}{\sqrt{d}}\right)V'\Bigr),
\end{equation}
where $Q$ is the query projected from the current spatial features, $(K,V)$ are the key and value projected from the textual context used in standard cross-attention, and $(K',V')$ are those from the H-Adapter conditioning features.
The output combines the standard cross-attention output with a masked H-Adapter conditioning branch, which is spatially gated by $M_t$ and scaled by $\lambda$.
The operator $\odot$ denotes element-wise multiplication.

\begin{figure}[t]
\centering
\begin{subfigure}{0.32\textwidth}
  \centering
  \includegraphics[width=\linewidth]{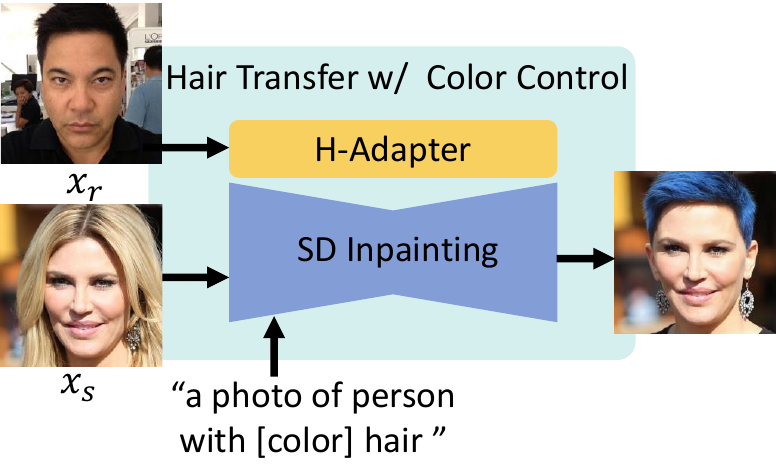}
  \caption{Auxiliary prompt-based control}
  \label{fig:method-overview-c}
\end{subfigure}\hfill
\begin{subfigure}{0.32\textwidth}
  \centering
  \includegraphics[width=\linewidth]{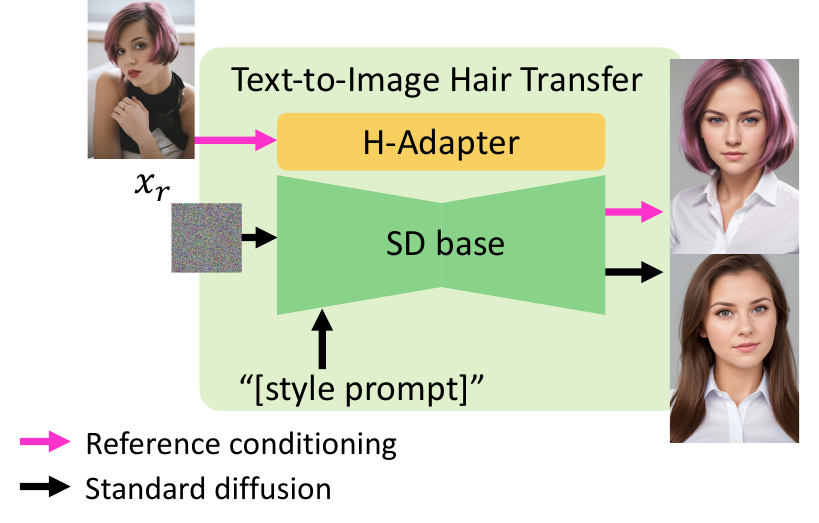}
  \caption{Source-free reference-guided text-to-image generation}
  \label{fig:method-overview-d}
\end{subfigure}\hfill
\begin{subfigure}{0.32\textwidth}
  \centering
  \includegraphics[width=\linewidth]{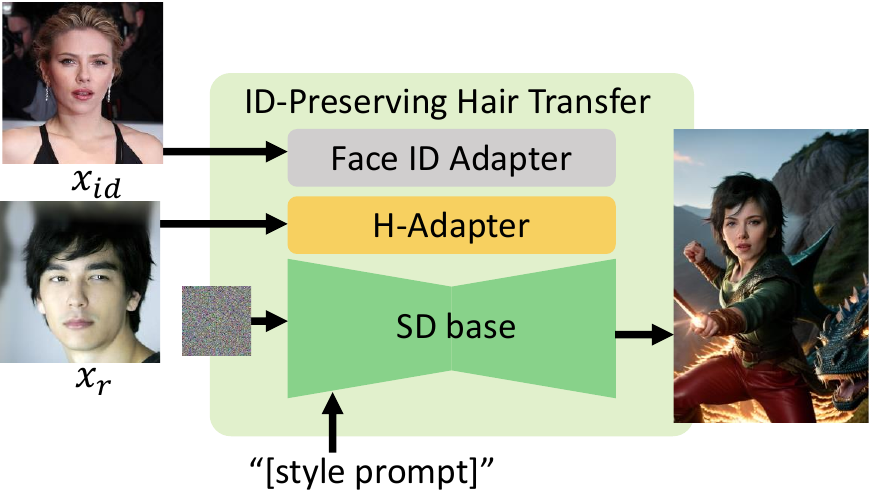}
  \caption{Identity-preserving hair transfer}
  \label{fig:method-overview-e}
\end{subfigure}

\caption{Flexible extensions of H-Adapter. H-Adapter supports auxiliary prompt control ($\lambda$; \cref{eq:method:ref-gated-attn}), source-free reference-guided text-to-image generation, and composition with identity-preserving adapters. Input images in panels (b) and (c) are adapted from Pexels stock images under the Pexels License and from \textcopyright{} JCS, Wikimedia Commons, licensed under CC BY 3.0, respectively.}
\label{fig:extensions}
\end{figure}

\subsubsection{Two-Stage Inpainting with a Warm-Up Step}

A two-stage inpainting pipeline with a warm-up step is employed.
\Cref{fig:method-overview-b} illustrates the overall procedure.
Since a source-aligned coarse mask is not available at the start of inference, we initialize inpainting with an all-one mask and run a single warm-up denoising step to extract cross-attention maps, from which the attention-based mask $M_{\mathrm{attn}}$ is derived.
Using $M_{\mathrm{attn}}$, denoising is run for a small number of steps (\eg, $\sim$20) to obtain an intermediate image $y_{\mathrm{mid}}$.
A pretrained segmentation model~\cite{yu2018bisenet} then predicts a pixel-level hair mask $M_{\mathrm{seg}}$ from the intermediate image $y_{\mathrm{mid}}$.
Denoising is then restarted from the initial noise latent $z_T$ with $M_{\mathrm{seg}}$ as the inpainting mask, yielding the final output $y_{\mathrm{fine}}$ with a pixel-level editable region.
Overall, this coarse-to-fine strategy progressively estimates the hair mask at increasing spatial precision, enabling robust source-pose-aligned hairstyle transfer.

\subsection{Extensions for Reference-Guided Hairstyle Transfer}
\label{sec:method:extensions}

As shown in \cref{fig:extensions}, H-Adapter supports flexible compositions with existing diffusion conditioning mechanisms.
We describe three practical usages built on the same interface.

\noindent\textbf{Auxiliary Prompt-Based Hair Color Control}
In the source-conditioned hairstyle transfer setting, we optionally incorporate an auxiliary text prompt (\eg, hair color) as an additional condition.
A relative weighting factor $\lambda$ (\cref{eq:method:ref-gated-attn}) controls the trade-off between auxiliary prompt controllability and reference faithfulness.

\noindent\textbf{Reference-guided text-to-image generation}
H-Adapter supports reference-guided text-to-image generation by injecting reference hairstyle features via cross-attention, without requiring a source-image condition.

\noindent\textbf{Compatibility with Identity-Preserving Adapters}
H-Adapter is composable with identity-preserving adapters by jointly conditioning the diffusion model on (i) a source identity embedding and (ii) the reference hairstyle embedding.
This enables identity-consistent hairstyle transfer while retaining prompt-driven content generation.

\section{Experiments}
\label{sec:exp}

\subsection{Experimental setup}
\label{sec:exp:experimental-setup}
H-Adapter is fine-tuned from IP-Adapter-Plus~\cite{ye2023ip} on Stable Diffusion v1.5~\cite{rombach2022high} using our region-specific loss.
We construct self-paired samples from FFHQ~\cite{karras2019style} images and same-identity video pairs from CelebV-HQ~\cite{zhu2022celebv} frames.
At inference time, H-Adapter is applied with the Stable Diffusion v1.5 inpainting model, using hair-removed base images synthesized by FLUX.2~\cite{bfl2026flux2klein}. Full implementation details are provided in Appendix~\ref{sup:detail} of the Supplementary Material.

\begin{figure}[tb]
  \centering
  \includegraphics[width=\linewidth]{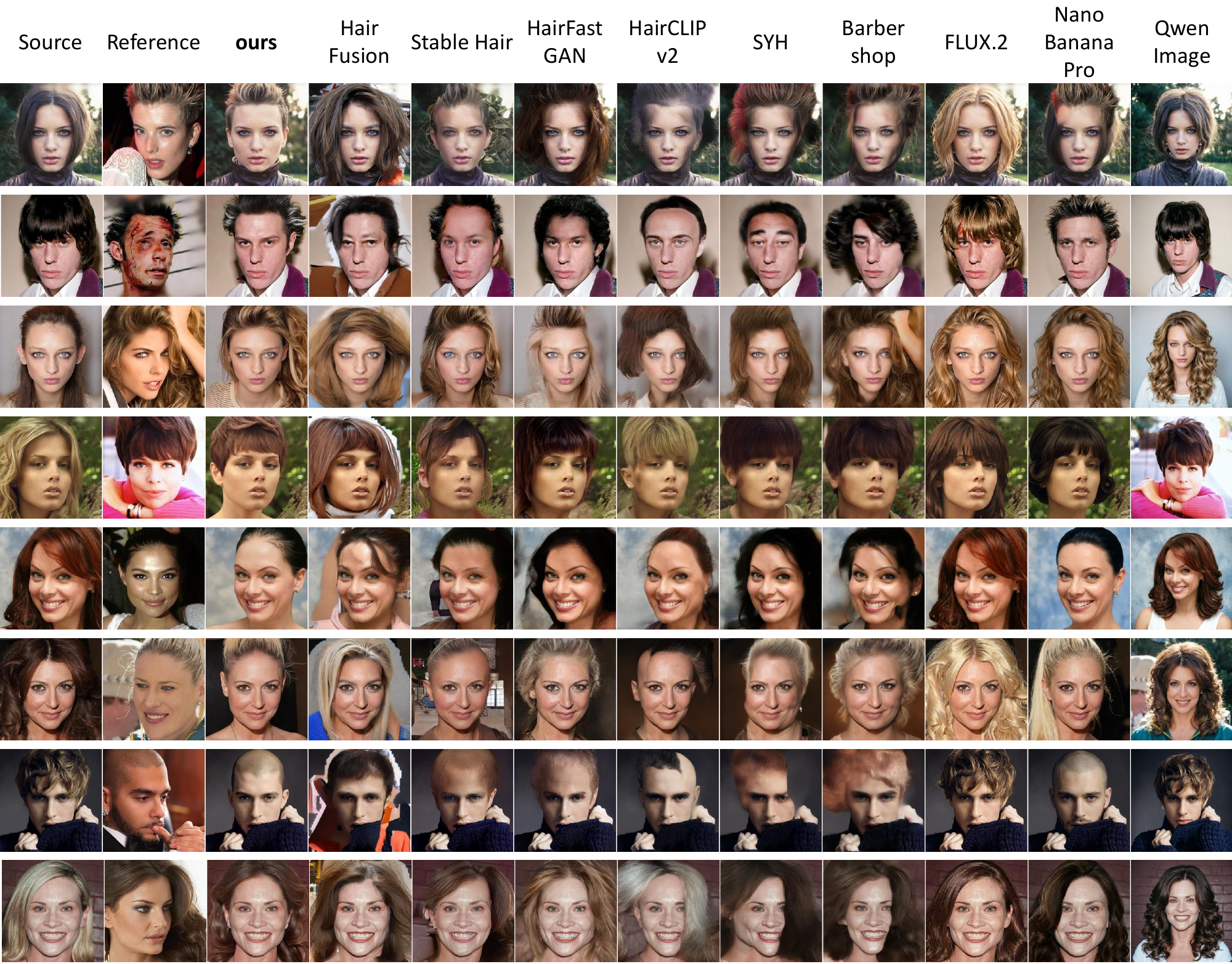}
  \caption{Qualitative comparisons under head-pose differences. The first two columns show the source and reference images; the remaining columns show outputs from each method. These examples span source--reference yaw gaps from $18.65^\circ$ to $53.21^\circ$.}
  \label{fig:qualitative-comparison}
\end{figure}

\subsection{Evaluation Protocol}
\label{sec:exp:eval-protocol}

\subsubsection{Baselines}
Comparisons are conducted against state-of-the-art hairstyle transfer methods, including HairCLIPv2~\cite{wei2023hairclipv2}, Style-Your-Hair (SYH)~\cite{kim2022style}, HairFastGAN~\cite{nikolaev2024hairfastgan}, Stable-Hair~\cite{zhang2025stable}, and HairFusion~\cite{chung2025preserve}. For each baseline, images are generated by following the official code release. We additionally report results obtained by replacing H-Adapter with the standard IP-Adapter~\cite{ye2023ip}, where the IP-Adapter is trained on the same dataset as ours for a fair comparison.
As an additional qualitative comparison, we further include recent general-purpose image editors---FLUX.2~\cite{bfl2026flux2klein}, Nano Banana Pro~\cite{deepmind2025nanobananapro}, and Qwen-Image-2.0-Pro~\cite{zhao2026qwen}---as well as the GAN-inversion-based method Barbershop~\cite{zhu2021barbershop}.

\begin{figure}[!t]
  \centering
  \includegraphics[width=\linewidth]{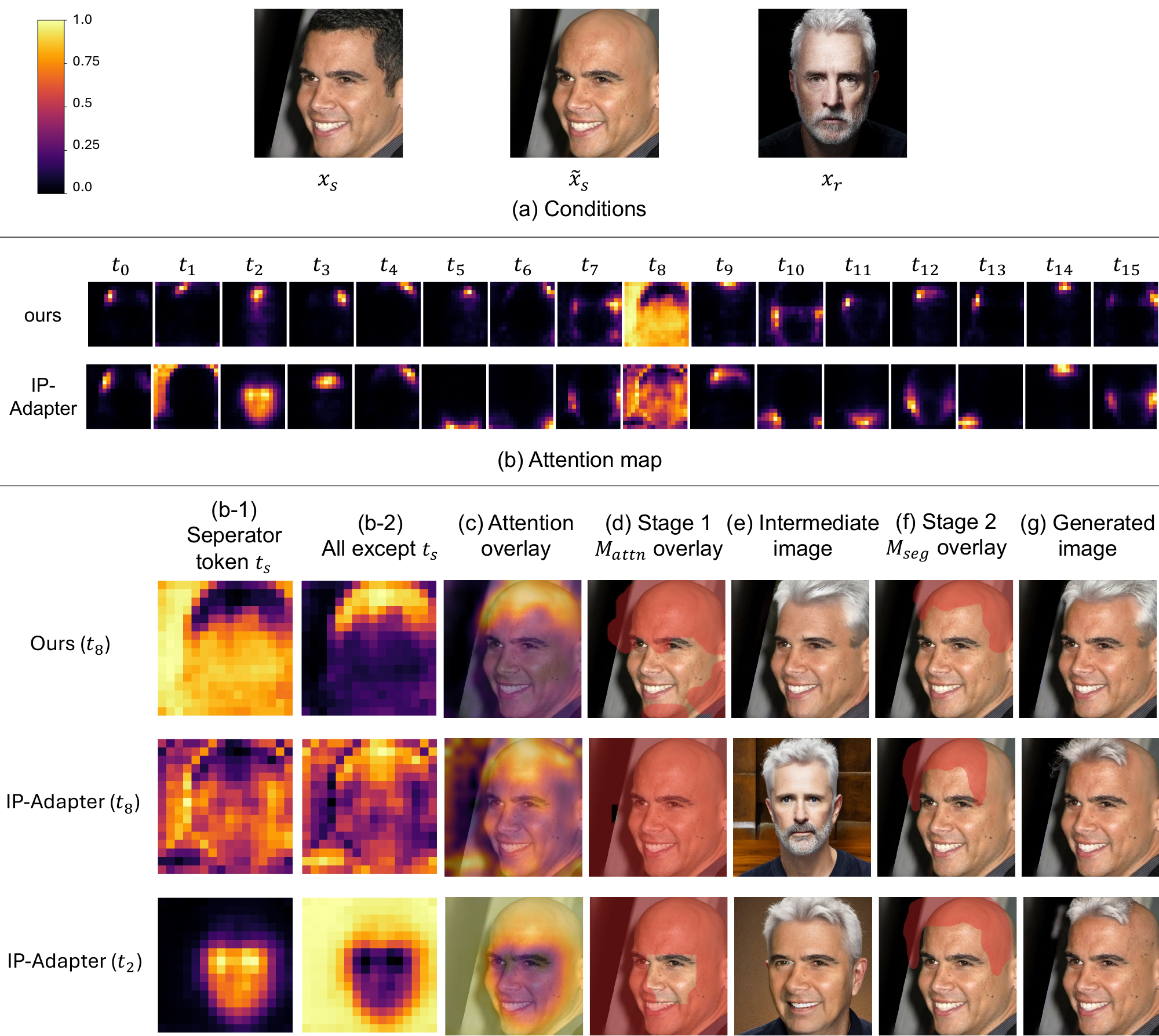}
  \caption{Stage-wise visualization of editing guidance in the proposed inference pipeline, comparing IP-Adapter and H-Adapter (ours). (a) Conditioning inputs: source $x_s$, hair-removed base $\tilde{x}_s$, and reference $x_r$. (b) Token-wise cross-attention for 16 tokens ${t_0,\ldots,t_{15}}$: (b-1) the attention map of the separator token $t_s$, with its token index shown in parentheses; and (b-2) the aggregated attention map obtained by summing all tokens except $t_s$. (c) Overlay of (b-2) on $\tilde{x}_s$. (d) Stage-1 inpainting mask $M_{\mathrm{attn}}$ overlaid on $\tilde{x}_s$. (e) Stage-1 intermediate output. (f) Stage-2 inpainting mask $M_{\mathrm{seg}}$ overlaid on $\tilde{x}_s$. (g) Stage-2 final output. The figure shows that region-specific loss yields more source-aligned guidance, reducing mask misalignment and improving pose- and shape-consistent hairstyle transfer.}
  \label{fig:exp:step_wise}
\end{figure}

\subsubsection{Metrics}
To evaluate overall distributional fidelity, we compute FID between the source images and the generated images.
$\mathrm{FID}_{\mathrm{CLIP}}$~\cite{kynkaanniemi2022role} is computed similarly to FID, but uses a CLIP image encoder~\cite{radford2021learning} instead of Inception~V3~\cite{szegedy2016rethinking}.
To evaluate non-hair preservation, reconstruction quality is measured on the intersection of the non-hair regions of the source and generated images using PSNR and SSIM~\cite{wang2004image}.
Hairstyle transfer faithfulness is measured by the CLIP-I~\cite{kynkaanniemi2022role} similarity between the reference image and the generated image.

\subsubsection{Evaluation subsets}
CelebA-HQ~\cite{karras2017progressive} contains 30{,}000 unpaired images.
We build two pair-disjoint evaluation subsets of 3{,}000 source--reference pairs by randomly pairing distinct images from the dataset, excluding hat images.
The pose-agnostic subset is sampled without conditioning on head pose, while the pose-different subset includes only pairs with an absolute yaw difference $>15^\circ$ to test robustness to pose variation.
For HairFusion~\cite{chung2025preserve}, facial landmark extraction fails for some samples; to ensure fair comparison, we remove the affected pairs (62 pose-agnostic, 101 pose-different) for all methods and report results on the remaining pairs.

\subsection{Qualitative Evaluation on Pose-Different Pairs}
\label{sec:exp:qualitative-evaluation}
\Cref{fig:qualitative-comparison} shows qualitative results on source–-reference pairs with different head poses, following the baseline setup described in \cref{sec:exp:eval-protocol}.
Our method transfers the reference hairstyle while aligning its shape and spatial placement to the source head geometry and pose, yielding coherent hair boundaries and natural integration with the source face.
In contrast, existing methods often fail to generate hair at an appropriate location or with a plausible shape aligned to the source head geometry, particularly with large differences in head shape or pose.
For general-purpose editors, the outputs can appear visually realistic, but the qualitative examples indicate that jointly preserving the source content and faithfully reflecting the reference hairstyle remains challenging under pose mismatch.

\begin{figure}[t]
  \centering
  \includegraphics[width=\linewidth]{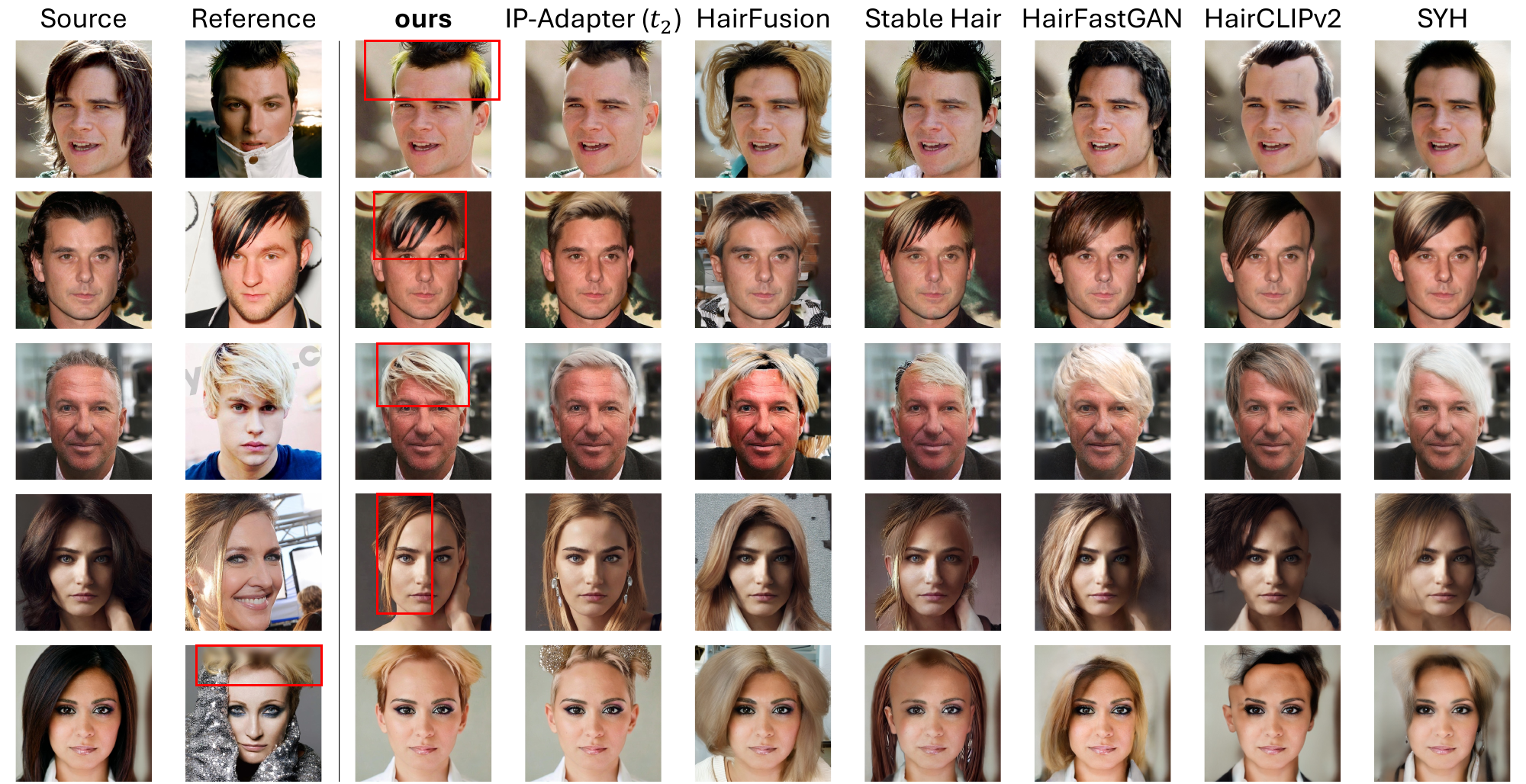}
  \caption{Qualitative analysis of reference-feature faithfulness. The first two columns show the source and reference conditions, while the remaining columns compare method outputs.}
\label{fig:reference_hairstyle_fidelity}
\end{figure}

\subsection{Analysis of Attention-Derived Localization}
\label{sec:exp:analysis}

We analyze the contribution of the region-specific loss to source-aligned mask generation in our two-stage inference pipeline.
To isolate its effect, we compare H-Adapter with an IP-Adapter baseline trained under the same setting but without the region-specific loss.
For the IP-Adapter baseline, we report both the same $t_8$-based extraction used in our pipeline and an alternative $t_2$-based variant to illustrate the instability of attention-based localization without the proposed objective.
As shown in \cref{fig:exp:step_wise}, H-Adapter, trained with the region-specific loss, produces a coarse but source-aligned attention mask in Stage~1, yielding an intermediate image from which Stage~2 extracts a more accurate pixel-level hair mask for localized inpainting.
In contrast, the IP-Adapter baselines exhibit less reliable attention localization, causing overly broad edits and boundary drift that degrade pose and shape consistency.
Detailed per-stage analyses are provided in Appendix~\ref{sup:stepwise} of the Supplementary Material.

\subsection{Qualitative Analysis of Reference Feature Faithfulness}
\label{sec:exp:qualitative-feature-faithfulness}
We compare reference hairstyle transfer fidelity across our method, existing methods, and an IP-Adapter baseline obtained by replacing our H-Adapter using the $t_2$-based mask.
As shown in \cref{fig:reference_hairstyle_fidelity}, our method faithfully transfers both fine-grained details and global hairstyle structure from the reference while remaining consistent with the source head geometry.
In the first two rows, it captures the color accents of the reference hairstyle.
The third row demonstrates improved transfer of fine-grained parting details.
The fourth row reproduces a thin highlight strand.
The last row highlights stable faithfulness under imperfect references: even when the reference hair region is blurred, our method preserves the intended hairstyle attributes.
Overall, these examples demonstrate faithful transfer of reference hairstyle attributes, preserving both fine-grained details and global structure.

\begin{table}[t]
\centering
\caption{Quantitative comparison on the pose-different subset.}
\label{tab:pd}
\begin{tabular}{lccccc}
\toprule
Method & FID $\downarrow$ & FID$_{\mathrm{CLIP}}$ $\downarrow$ & SSIM $\uparrow$ & PSNR $\uparrow$ & CLIP-I $\uparrow$ \\
\midrule
Ours                       & \textbf{12.47} & \textbf{3.98} & \textbf{0.831} & 23.06 & \textbf{0.659} \\
\midrule
IP-Adapter ($t_8$)          & 15.27 & 8.83 & 0.803 & 21.56 & 0.639 \\
IP-Adapter ($t_2$)          & 12.53 & 4.26 & 0.825 & 22.70 & 0.651 \\
\midrule
HairFusion                 & 28.03 & 8.80 & 0.756 & 17.26 & 0.626 \\
Stable-Hair                & 25.79 & 8.70 & 0.798 & 22.39 & 0.640 \\
HairFastGAN                & 12.78 & 4.53 & 0.817 & 24.40 & 0.649 \\
HairCLIPv2                 & 13.44 & 7.91 & 0.824 & 23.63 & 0.623 \\
Style-Your-Hair            & 15.95 & 8.54 & 0.816 & 22.80 & 0.649 \\
\bottomrule
\end{tabular}
\end{table}

\subsection{Quantitative Evaluation}

\Cref{tab:pd} reports quantitative results on the pose-different subset, while results on the pose-agnostic subset are provided in Appendix~\ref{sup:quant} of the Supplementary Material.
On the pose-different subset, our method outperforms prior approaches as well as an IP-Adapter baseline obtained by replacing our H-Adapter, achieving the best FID and FID$_{\mathrm{CLIP}}$ scores, indicating improved visual fidelity.
Notably, it attains the highest CLIP-I, suggesting the most faithful transfer of reference hairstyle attributes under pose mismatch.
Moreover, it maintains high SSIM with competitive PSNR, indicating that structural similarity outside the edited hair region is largely preserved.

\begin{table}[t]
\centering
\caption{VLM-as-a-judge scores (GPT-4o) with 95\% bootstrap percentile CIs ($n_{\text{boot}}=1{,}000$ over per-triplet template-mean scores).}
\label{tab:mean_scores_combined}
\begin{tabular}{lccc}
\toprule
Method & HFS$\uparrow$ & NPS$\uparrow$ & AQS$\uparrow$ \\
\midrule
Ours            & \textbf{3.11}~{\scriptsize[2.99, 3.23]} & \textbf{4.23}~{\scriptsize[4.15, 4.32]} & \textbf{3.73}~{\scriptsize[3.63, 3.83]} \\
\midrule
HairFusion      & 2.55~{\scriptsize[2.44, 2.65]} & 3.55~{\scriptsize[3.46, 3.64]} & 3.09~{\scriptsize[2.99, 3.18]} \\
Stable-Hair     & 2.90~{\scriptsize[2.77, 3.02]} & 3.42~{\scriptsize[3.33, 3.51]} & 2.75~{\scriptsize[2.67, 2.84]} \\
HairFastGAN     & 2.83~{\scriptsize[2.72, 2.95]} & 3.91~{\scriptsize[3.82, 4.00]} & 3.29~{\scriptsize[3.19, 3.38]} \\
HairCLIPv2      & 2.10~{\scriptsize[1.97, 2.23]} & 4.05~{\scriptsize[3.95, 4.13]} & 3.57~{\scriptsize[3.46, 3.67]} \\
Style-Your-Hair & 2.87~{\scriptsize[2.74, 3.02]} & 3.98~{\scriptsize[3.90, 4.07]} & 3.61~{\scriptsize[3.52, 3.70]} \\
\bottomrule
\end{tabular}
\end{table}

\begin{table}[t]
\centering
\caption{Human preference study results, including valid pairwise votes, H-Adapter preference rates, 95\% confidence intervals, and two-sided exact binomial test $p$-values against a 50\% chance level.}
\label{tab:user_study}
\begin{tabular}{lcccc}
\hline
Baseline & Votes & Ours (\%) & 95\% CI & $p$-value \\
\hline
HairCLIPv2 & 624 & 81.1 & [77.8, 84.0] & $<0.001$ \\
HairFastGAN & 595 & 55.3 & [51.3, 59.2] & 0.011 \\
HairFusion & 678 & 80.4 & [77.2, 83.2] & $<0.001$ \\
Stable-Hair & 656 & 72.9 & [69.3, 76.1] & $<0.001$ \\
Style-Your-Hair & 640 & 72.3 & [68.8, 75.7] & $<0.001$ \\
\hline
Overall & 3193 & 72.7 & [71.1, 74.2]  & $<0.001$ \\
\hline
\end{tabular}
\end{table}

\subsection{VLM-as-a-Judge Evaluation}
\label{sec:vlm_judge}

While the above metrics quantify overall quality and pixel-level non-hair consistency, they cannot isolate hair-specific transfer quality, semantic non-hair preservation, or localized artifacts.
We therefore adopt a VLM-as-a-judge framework~\cite{zhang2023gpt} to score each result along three axes:
(i)~\textsc{Hair Fidelity Score (\texttt{HFS})}---how faithfully the reference hairstyle is reproduced;
(ii)~\textsc{Non-hair Preservation Score (\texttt{NPS})}---how well identity, background, and other non-hair regions are preserved; and
(iii)~\textsc{Artifact Quality Score (\texttt{AQS})}---the absence of artifacts such as blending seams, color bleeding, or unnatural boundaries.

\subsubsection{Evaluation Protocol.}

For each model, we evaluate 200 triplets (source, reference, generated output). The source--reference pairs are randomly sampled from CelebA-HQ and shared across all models.
We adopt a \emph{pointwise} scoring protocol~\cite{chen2024mllm} in which each VLM judge assigns a 1–5 Likert score to each axis via separate prompts to avoid inter-axis interference~\cite{ku2024viescore}.
To mitigate prompt-sensitivity bias~\cite{slyman2025calibrating}, we construct three lexically distinct prompt variants per axis and average across templates in each axis.
Each prompt incorporates rubric anchors~\cite{lee2024prometheus} and chain-of-thought rationale elicitation~\cite{ku2024viescore}.
To reduce single-judge bias, the pipeline is executed with three VLM judges---GPT-4o~\cite{hurst2024gpt4o}, GPT-5.2~\cite{openai2025gpt5, openai2025gpt52}, and Gemini-2.5-Flash~\cite{comanici2025gemini25}---with prompt templates, consistency and pairwise validation reported in Appendix~\ref{sup:vlm} of the Supplementary Material.

\subsubsection{Evaluation Results.}
\label{sec:results}

\begin{figure}[t]
    \centering
    \includegraphics[width=\linewidth]{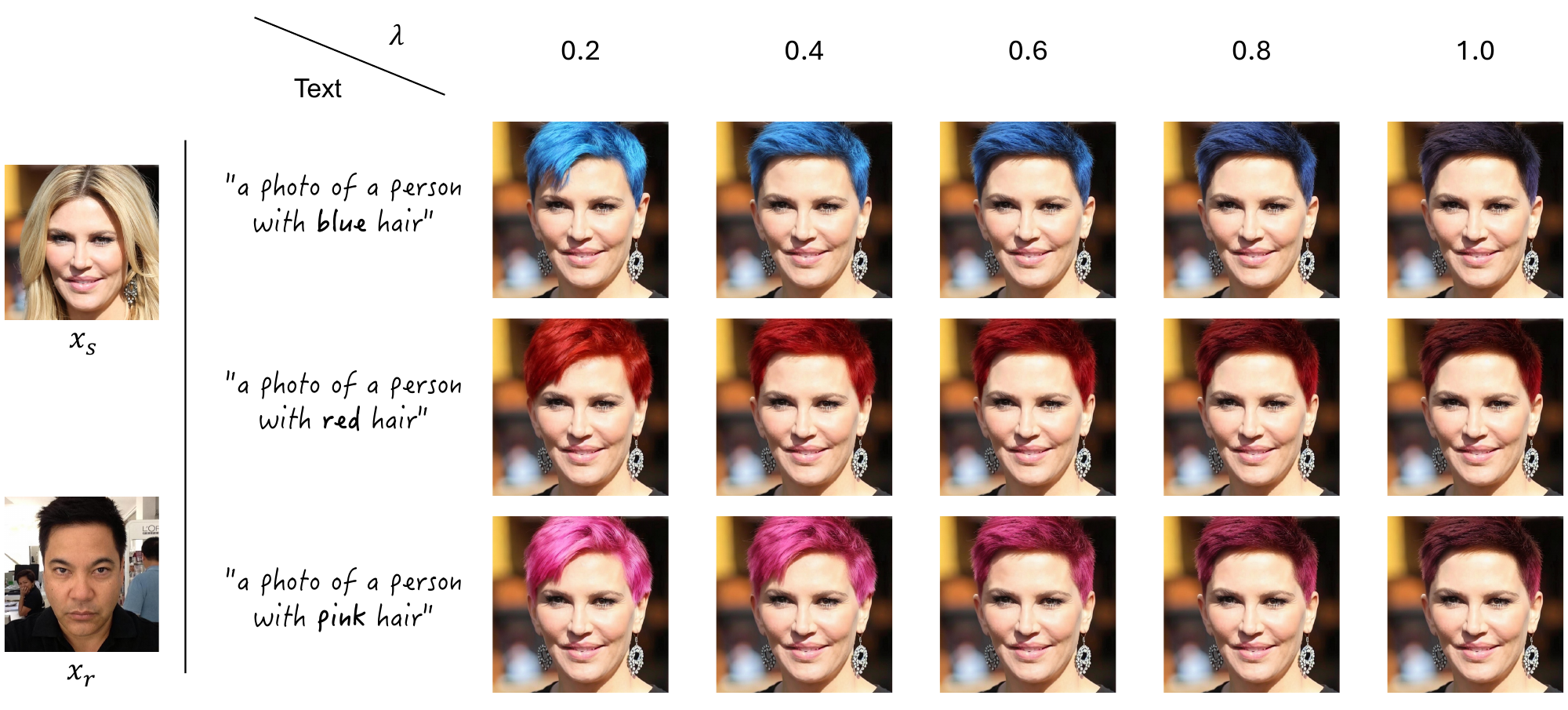}
    \caption{Auxiliary hair-color control with H-Adapter. Inputs and the random seed are fixed. Rows vary hair-color prompts and columns vary $\lambda$ ($0.2$–$1.0$). Smaller $\lambda$ improves prompt responsiveness, while larger $\lambda$ strengthens reference faithfulness.}
    \label{fig:exp:prompt-control}
\end{figure}

We present results with GPT-4o as the representative judge, as it achieves the highest consistency (Krippendorff's $\alpha \geq 0.90$ on all axes; mean per-triplet run-to-run $\sigma \leq 0.07$).

As shown in \cref{tab:mean_scores_combined}, our method achieves the highest mean on all three axes.
While the AQS gap over Style-Your-Hair falls within overlapping confidence intervals, Style-Your-Hair ranks third on HFS.
Other baselines show similar trade-offs (\eg, HairCLIPv2 is second on NPS but last on HFS), whereas ours is the only method that maintains top-tier performance across all criteria without such compromises.
Note that HFS scores are generally low across all methods due to the strict rubric requiring simultaneous match in color, texture, length, silhouette, and parting; ours is the only method exceeding 3.0.
Pairwise validation and per-judge breakdowns are provided in Appendix~\ref{sup:vlm} of the Supplementary Material.

\subsection{Human Preference Study}
To complement automatic and VLM-based evaluations, we conduct a two-alternative forced-choice (2AFC) human preference study. Given the same source and reference images, participants select the result they preferred
overall between two anonymized outputs, one from H-Adapter and one from a baseline. Participants may answer an arbitrary number of questions. As shown in \cref{tab:user_study}, we report 3{,}193 valid pairwise votes from 53 respondents. Our method is preferred in 72.7\% of comparisons (binomial test, $p < 0.001$) and is favored over every baseline individually.

\begin{figure}[t]
    \centering
    \includegraphics[width=\linewidth]{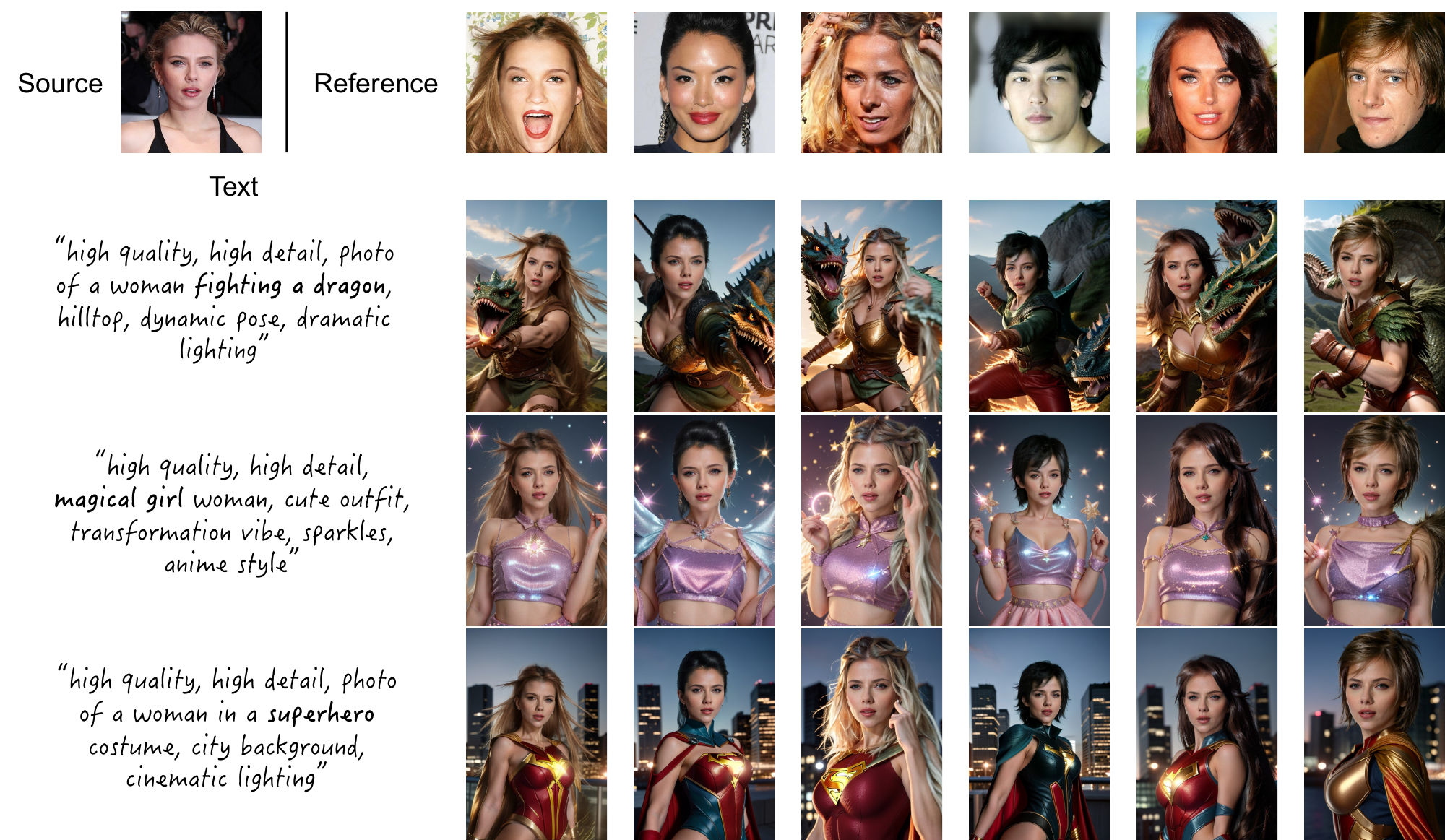}
    \caption{Qualitative results of H-Adapter combined with IP-Adapter FaceID Plus~\cite{ye2023ip}. The source image conditions FaceID Plus for identity preservation. Rows vary text prompts, and columns vary reference images for H-Adapter, demonstrating reference-guided hair transfer across diverse contexts while preserving identity. The source image is adapted from \textcopyright{} JCS, Wikimedia Commons, licensed under CC BY 3.0.}
    \label{fig:exp:personalization}
\end{figure}

\subsection{Flexible Usage of H-Adapter}
Beyond the default hairstyle transfer setting, H-Adapter also supports auxiliary prompt-based hair-color control, reference-guided text-to-image generation, and composition with IP-Adapter FaceID Plus for identity-preserving generation. As shown in \cref{fig:exp:prompt-control,fig:exp:personalization}, these usages demonstrate that the learned hair-conditioning branch can be flexibly combined with text prompts and identity conditions without retraining. Additional qualitative results, including in-the-wild examples, stylized inputs, reference-guided text-to-image generation, and identity-preserving adapter analysis, are provided in Appendix~\ref{sup:results} of the Supplementary Material.



\section{Conclusion}
\label{sec:conclusion}

We present H-Adapter, a pose-robust hairstyle transfer framework that trains an IP-Adapter with a region-specific objective to disentangle hair and non-hair regions. The resulting cross-attention provides a source-aligned coarse mask, which guides a two-stage diffusion inpainting pipeline to align the transferred hairstyle with the source head geometry. Across quantitative, qualitative, VLM-as-a-judge, and human preference evaluations, H-Adapter improves reference faithfulness and non-hair preservation under pose mismatch, while supporting flexible uses such as prompt-based hair-color control and identity-preserving generation. More broadly, our results suggest that region-specific training can turn cross-attention into a useful localization signal in reference-guided local editing.





%
%
\bibliographystyle{splncs04}
\bibliography{main}

@String(ECCV  = {Eur. Conf. Comput. Vis.})

@String(AAAI  = {AAAI})

@String(ECCV  = {ECCV})

@online{ipadapter,
  author  = {{h94}},
  title   = {IP-Adapter},
  url     = {https://huggingface.co/h94/IP-Adapter},
  urldate = {2026-03-10},
  note = {Accessed: 2026-03-10}
}

@online{ipadapter_faceid,
  author  = {{h94}},
  title   = {IP-Adapter-FaceID},
  url     = {https://huggingface.co/h94/IP-Adapter-FaceID},
  urldate = {2026-03-10},
  note = {Accessed: 2026-03-10}
}

@online{clip_vit_h_14,
  author  = {{LAION}},
  title   = {CLIP-ViT-H-14-laion2B-s32B-b79K},
  url     = {https://huggingface.co/laion/CLIP-ViT-H-14-laion2B-s32B-b79K},
  urldate = {2026-03-10},
  note = {Accessed: 2026-03-10}
}

@online{sd15_inpainting,
  author  = {{stabilityai}},
  title   = {stable-diffusion-v1-5/stable-diffusion-inpainting},
  url     = {https://huggingface.co/stable-diffusion-v1-5/stable-diffusion-inpainting},
  urldate = {2026-03-10},
  note = {Accessed: 2026-03-10}
}

@online{realistic_vision_v4_novae,
  author  = {{SG161222}},
  title   = {Realistic Vision V4.0 noVAE},
  url     = {https://huggingface.co/SG161222/Realistic_Vision_V4.0_noVAE},
  urldate = {2026-03-10},
  note = {Accessed: 2026-03-10}
}

@online{sd_vae_ft_mse,
  author  = {{Stability AI}},
  title   = {sd-vae-ft-mse},
  url     = {https://huggingface.co/stabilityai/sd-vae-ft-mse},
  urldate = {2026-03-10},
  note = {Accessed: 2026-03-10}
}

@inproceedings{yu2018bisenet,
  title={Bisenet: Bilateral segmentation network for real-time semantic segmentation},
  author={Yu, Changqian and Wang, Jingbo and Peng, Chao and Gao, Changxin and Yu, Gang and Sang, Nong},
  booktitle={Proceedings of the European conference on computer vision (ECCV)},
  pages={325--341},
  year={2018}
}

@article{kvanchiani2023easyportrait,
  title={EasyPortrait--Face Parsing and Portrait Segmentation Dataset},
  author={Kvanchiani, Karina and Petrova, Elizaveta and Efremyan, Karen and Sautin, Alexander and Kapitanov, Alexander},
  journal={arXiv preprint arXiv:2304.13509},
  year={2023}
}

@article{wang2004image,
  title={Image quality assessment: from error visibility to structural similarity},
  author={Wang, Zhou and Bovik, Alan C and Sheikh, Hamid R and Simoncelli, Eero P},
  journal={IEEE transactions on image processing},
  volume={13},
  number={4},
  pages={600--612},
  year={2004},
  publisher={IEEE}
}

@inproceedings{szegedy2016rethinking,
  title={Rethinking the inception architecture for computer vision},
  author={Szegedy, Christian and Vanhoucke, Vincent and Ioffe, Sergey and Shlens, Jon and Wojna, Zbigniew},
  booktitle={Proceedings of the IEEE conference on computer vision and pattern recognition},
  pages={2818--2826},
  year={2016}
}

@inproceedings{radford2021learning,
  title={Learning transferable visual models from natural language supervision},
  author={Radford, Alec and Kim, Jong Wook and Hallacy, Chris and Ramesh, Aditya and Goh, Gabriel and Agarwal, Sandhini and Sastry, Girish and Askell, Amanda and Mishkin, Pamela and Clark, Jack and others},
  booktitle={International conference on machine learning},
  pages={8748--8763},
  year={2021},
  organization={PmLR}
}

@inproceedings{zhu2022celebv,
  title={CelebV-HQ: A large-scale video facial attributes dataset},
  author={Zhu, Hao and Wu, Wayne and Zhu, Wentao and Jiang, Liming and Tang, Siwei and Zhang, Li and Liu, Ziwei and Loy, Chen Change},
  booktitle={European conference on computer vision},
  pages={650--667},
  year={2022},
  organization={Springer}
}

@inproceedings{karras2019style,
  title={A style-based generator architecture for generative adversarial networks},
  author={Karras, Tero and Laine, Samuli and Aila, Timo},
  booktitle={Proceedings of the IEEE/CVF conference on computer vision and pattern recognition},
  pages={4401--4410},
  year={2019}
}

@article{karras2017progressive,
  title={Progressive growing of gans for improved quality, stability, and variation},
  author={Karras, Tero and Aila, Timo and Laine, Samuli and Lehtinen, Jaakko},
  journal={arXiv preprint arXiv:1710.10196},
  year={2017}
}

@article{ye2023ip,
  title={Ip-adapter: Text compatible image prompt adapter for text-to-image diffusion models},
  author={Ye, Hu and Zhang, Jun and Liu, Sibo and Han, Xiao and Yang, Wei},
  journal={arXiv preprint arXiv:2308.06721},
  year={2023}
}

@inproceedings{guler2018densepose,
  title={Densepose: Dense human pose estimation in the wild},
  author={G{\"u}ler, R{\i}za Alp and Neverova, Natalia and Kokkinos, Iasonas},
  booktitle={Proceedings of the IEEE conference on computer vision and pattern recognition},
  pages={7297--7306},
  year={2018}
}

@inproceedings{rombach2022high,
  title={High-resolution image synthesis with latent diffusion models},
  author={Rombach, Robin and Blattmann, Andreas and Lorenz, Dominik and Esser, Patrick and Ommer, Bj{\"o}rn},
  booktitle={Proceedings of the IEEE/CVF conference on computer vision and pattern recognition},
  pages={10684--10695},
  year={2022}
}

@article{saharia2022photorealistic,
  title={Photorealistic text-to-image diffusion models with deep language understanding},
  author={Saharia, Chitwan and Chan, William and Saxena, Saurabh and Li, Lala and Whang, Jay and Denton, Emily L and Ghasemipour, Kamyar and Gontijo Lopes, Raphael and Karagol Ayan, Burcu and Salimans, Tim and others},
  journal={Advances in neural information processing systems},
  volume={35},
  pages={36479--36494},
  year={2022}
}

@article{podell2023sdxl,
  title={Sdxl: Improving latent diffusion models for high-resolution image synthesis},
  author={Podell, Dustin and English, Zion and Lacey, Kyle and Blattmann, Andreas and Dockhorn, Tim and M{\"u}ller, Jonas and Penna, Joe and Rombach, Robin},
  journal={arXiv preprint arXiv:2307.01952},
  year={2023}
}

@article{nichol2021glide,
  title={Glide: Towards photorealistic image generation and editing with text-guided diffusion models},
  author={Nichol, Alex and Dhariwal, Prafulla and Ramesh, Aditya and Shyam, Pranav and Mishkin, Pamela and McGrew, Bob and Sutskever, Ilya and Chen, Mark},
  journal={arXiv preprint arXiv:2112.10741},
  year={2021}
}

@article{ramesh2022hierarchical,
  title={Hierarchical text-conditional image generation with clip latents},
  author={Ramesh, Aditya and Dhariwal, Prafulla and Nichol, Alex and Chu, Casey and Chen, Mark},
  journal={arXiv preprint arXiv:2204.06125},
  volume={1},
  number={2},
  pages={3},
  year={2022}
}

@online{bfl2026flux2klein,
  author   = {{Black Forest Labs}},
  title    = {FLUX.2 [klein]: Towards Interactive Visual Intelligence},
  date     = {2026-01-15},
  url      = {https://bfl.ai/blog/flux2-klein-towards-interactive-visual-intelligence},
  urldate  = {2026-03-05},
  note         = {Accessed: 2026-03-05}
}

@online{deepmind2025nanobananapro,
  author       = {{Google DeepMind}},
  title        = {Gemini 3 Pro Image -- Nano Banana Pro},
  url          = {https://deepmind.google/models/gemini-image/pro/},
  urldate      = {2026-06-22},
  organization = {Google DeepMind},
  note         = {Accessed: 2026-06-22}
}

@article{zhao2026qwen,
  title={Qwen-Image-2.0 Technical Report},
  author={Zhao, Bing and Wu, Chenfei and Li, Deqing and Meng, Hao and Li, Jiahao and Zhang, Jie and Zhou, Jingren and Lin, Junyang and Gao, Kaiyuan and Cao, Kuan and others},
  journal={arXiv preprint arXiv:2605.10730},
  year={2026}
}

@article{kynkaanniemi2022role,
  title={The role of imagenet classes in fr$\backslash$'echet inception distance},
  author={Kynk{\"a}{\"a}nniemi, Tuomas and Karras, Tero and Aittala, Miika and Aila, Timo and Lehtinen, Jaakko},
  journal={arXiv preprint arXiv:2203.06026},
  year={2022}
}

@article{zeng2024hairdiffusion,
  title={HairDiffusion: Vivid multi-colored hair editing via latent diffusion},
  author={Zeng, Yu and Zhang, Yang and Jiachen, Liu and Shen, Linlin and Deng, Kaijun and He, Weizhao and Wang, Jinbao},
  journal={Advances in Neural Information Processing Systems},
  volume={37},
  pages={5048--5073},
  year={2024}
}

@inproceedings{zhang2025stable,
  title={Stable-hair: Real-world hair transfer via diffusion model},
  author={Zhang, Yuxuan and Zhang, Qing and Song, Yiren and Zhang, Jichao and Tang, Hao and Liu, Jiaming},
  booktitle={Proceedings of the AAAI Conference on Artificial Intelligence},
  volume={39},
  number={10},
  pages={10348--10356},
  year={2025}
}

@inproceedings{chung2025preserve,
  title={What to preserve and what to transfer: Faithful, identity-preserving diffusion-based hairstyle transfer},
  author={Chung, Chaeyeon and Park, Sunghyun and Kim, Jeongho and Choo, Jaegul},
  booktitle={Proceedings of the AAAI Conference on Artificial Intelligence},
  volume={39},
  number={3},
  pages={2582--2590},
  year={2025}
}

@inproceedings{saha2021loho,
  title={Loho: Latent optimization of hairstyles via orthogonalization},
  author={Saha, Rohit and Duke, Brendan and Shkurti, Florian and Taylor, Graham W and Aarabi, Parham},
  booktitle={Proceedings of the IEEE/CVF Conference on Computer Vision and Pattern Recognition},
  pages={1984--1993},
  year={2021}
}

@article{tan2020michigan,
  title={Michigan: multi-input-conditioned hair image generation for portrait editing},
  author={Tan, Zhentao and Chai, Menglei and Chen, Dongdong and Liao, Jing and Chu, Qi and Yuan, Lu and Tulyakov, Sergey and Yu, Nenghai},
  journal={arXiv preprint arXiv:2010.16417},
  year={2020}
}

@article{zhu2021barbershop,
  title={Barbershop: Gan-based image compositing using segmentation masks},
  author={Zhu, Peihao and Abdal, Rameen and Femiani, John and Wonka, Peter},
  journal={arXiv preprint arXiv:2106.01505},
  year={2021}
}

@inproceedings{wei2022hairclip,
  title={Hairclip: Design your hair by text and reference image},
  author={Wei, Tianyi and Chen, Dongdong and Zhou, Wenbo and Liao, Jing and Tan, Zhentao and Yuan, Lu and Zhang, Weiming and Yu, Nenghai},
  booktitle={Proceedings of the IEEE/CVF conference on computer vision and pattern recognition},
  pages={18072--18081},
  year={2022}
}

@inproceedings{wei2023hairclipv2,
  title={Hairclipv2: Unifying hair editing via proxy feature blending},
  author={Wei, Tianyi and Chen, Dongdong and Zhou, Wenbo and Liao, Jing and Zhang, Weiming and Hua, Gang and Yu, Nenghai},
  booktitle={Proceedings of the IEEE/CVF International Conference on Computer Vision},
  pages={23589--23599},
  year={2023}
}

@article{wei2026unifying,
  title={Unifying Multi-modal Hair Editing via Proxy Feature Blending},
  author={Wei, Tianyi and Chen, Dongdong and Zhou, Wenbo and Liao, Jing and Wang, Can and Zhang, Weiming and Hua, Gang and Yu, Nenghai},
  journal={IEEE Transactions on Pattern Analysis and Machine Intelligence},
  year={2026},
  publisher={IEEE}
}

@article{nikolaev2024hairfastgan,
  title={Hairfastgan: Realistic and robust hair transfer with a fast encoder-based approach},
  author={Nikolaev, Maxim and Kuznetsov, Mikhail and Vetrov, Dmitry and Alanov, Aibek},
  journal={Advances in Neural Information Processing Systems},
  volume={37},
  pages={45600--45635},
  year={2024}
}

@inproceedings{kim2022style,
  title={Style your hair: Latent optimization for pose-invariant hairstyle transfer via local-style-aware hair alignment},
  author={Kim, Taewoo and Chung, Chaeyeon and Kim, Yoonseo and Park, Sunghyun and Kim, Kangyeol and Choo, Jaegul},
  booktitle={European conference on computer vision},
  pages={188--203},
  year={2022},
  organization={Springer}
}

@article{chung2022hairfit,
  title={Hairfit: pose-invariant hairstyle transfer via flow-based hair alignment and semantic-region-aware inpainting},
  author={Chung, Chaeyeon and Kim, Taewoo and Nam, Hyelin and Choi, Seunghwan and Gu, Gyojung and Park, Sunghyun and Choo, Jaegul},
  journal={arXiv preprint arXiv:2206.08585},
  year={2022}
}

@inproceedings{zhu2022hairnet,
  title={Hairnet: Hairstyle transfer with pose changes},
  author={Zhu, Peihao and Abdal, Rameen and Femiani, John and Wonka, Peter},
  booktitle={European Conference on Computer Vision},
  pages={651--667},
  year={2022},
  organization={Springer}
}

@inproceedings{cao2023masactrl,
  title={Masactrl: Tuning-free mutual self-attention control for consistent image synthesis and editing},
  author={Cao, Mingdeng and Wang, Xintao and Qi, Zhongang and Shan, Ying and Qie, Xiaohu and Zheng, Yinqiang},
  booktitle={Proceedings of the IEEE/CVF international conference on computer vision},
  pages={22560--22570},
  year={2023}
}

@article{couairon2022diffedit,
  title={Diffedit: Diffusion-based semantic image editing with mask guidance},
  author={Couairon, Guillaume and Verbeek, Jakob and Schwenk, Holger and Cord, Matthieu},
  journal={arXiv preprint arXiv:2210.11427},
  year={2022}
}

@inproceedings{zou2024towards,
  title={Towards efficient diffusion-based image editing with instant attention masks},
  author={Zou, Siyu and Tang, Jiji and Zhou, Yiyi and He, Jing and Zhao, Chaoyi and Zhang, Rongsheng and Hu, Zhipeng and Sun, Xiaoshuai},
  booktitle={Proceedings of the AAAI Conference on Artificial Intelligence},
  volume={38},
  number={7},
  pages={7864--7872},
  year={2024}
}

@article{hertz2022prompt,
  title={Prompt-to-prompt image editing with cross attention control},
  author={Hertz, Amir and Mokady, Ron and Tenenbaum, Jay and Aberman, Kfir and Pritch, Yael and Cohen-Or, Daniel},
  journal={arXiv preprint arXiv:2208.01626},
  year={2022}
}

@inproceedings{tumanyan2023plug,
  title={Plug-and-play diffusion features for text-driven image-to-image translation},
  author={Tumanyan, Narek and Geyer, Michal and Bagon, Shai and Dekel, Tali},
  booktitle={Proceedings of the IEEE/CVF conference on computer vision and pattern recognition},
  pages={1921--1930},
  year={2023}
}

@article{zhang2023gpt,
  title={Gpt-4v (ision) as a generalist evaluator for vision-language tasks},
  author={Zhang, Xinlu and Lu, Yujie and Wang, Weizhi and Yan, An and Yan, Jun and Qin, Lianke and Wang, Heng and Yan, Xifeng and Wang, William Yang and Petzold, Linda Ruth},
  journal={arXiv preprint arXiv:2311.01361},
  year={2023}
}

@inproceedings{chen2024mllm,
  title={Mllm-as-a-judge: Assessing multimodal llm-as-a-judge with vision-language benchmark},
  author={Chen, Dongping and Chen, Ruoxi and Zhang, Shilin and Wang, Yaochen and Liu, Yinuo and Zhou, Huichi and Zhang, Qihui and Wan, Yao and Zhou, Pan and Sun, Lichao},
  booktitle={Forty-first International Conference on Machine Learning},
  year={2024}
}

@inproceedings{lee2024prometheus,
  title={Prometheus-vision: Vision-language model as a judge for fine-grained evaluation},
  author={Lee, Seongyun and Kim, Seungone and Park, Sue and Kim, Geewook and Seo, Minjoon},
  booktitle={Findings of the Association for Computational Linguistics: ACL 2024},
  pages={11286--11315},
  year={2024}
}

@inproceedings{ku2024viescore,
  title={Viescore: Towards explainable metrics for conditional image synthesis evaluation},
  author={Ku, Max and Jiang, Dongfu and Wei, Cong and Yue, Xiang and Chen, Wenhu},
  booktitle={Proceedings of the 62nd Annual Meeting of the Association for Computational Linguistics (Volume 1: Long Papers)},
  pages={12268--12290},
  year={2024}
}

@inproceedings{slyman2025calibrating,
  title={Calibrating MLLM-as-a-judge via Multimodal Bayesian Prompt Ensembles},
  author={Slyman, Eric and Tanjim, Mehrab and Kafle, Kushal and Lee, Stefan},
  booktitle={Proceedings of the IEEE/CVF International Conference on Computer Vision},
  pages={17224--17234},
  year={2025}
}

@article{hurst2024gpt4o,
  title={GPT-4o System Card},
  author={OpenAI},
  journal={arXiv preprint arXiv:2410.21276},
  year={2024}
}

@article{openai2025gpt5,
  title={OpenAI GPT-5 System Card},
  author={OpenAI},
  journal={arXiv preprint arXiv:2601.03267},
  year={2025}
}

@misc{openai2025gpt52,
  title={Update to GPT-5 System Card: GPT-5.2},
  author={OpenAI},
  year={2025},
  howpublished={\url{https://openai.com/index/gpt-5-system-card-update-gpt-5-2/}},
  note={Acessed: 2026-06-30}
}

@article{comanici2025gemini25,
  title={Gemini 2.5: Pushing the Frontier with Advanced Reasoning, 
         Multimodality, Long Context, and Next Generation Agentic Capabilities},
  author={Comanici, Gheorghe and others},
  journal={arXiv preprint arXiv:2507.06261},
  year={2025}
}
\clearpage

\appendix
\section*{Supplementary Material}

\section{Stage-wise Analysis of Source-Aligned Mask Generation}
\label{sup:stepwise}

This section provides a detailed analysis of the stage-wise visualization in \cref{fig:exp:step_wise} of the main paper.
For our method, the attention-derived localization is already source-aligned when overlaid on the bald input in panels (b--c).
Although the resulting $M_{\mathrm{attn}}$ in panel (d) remains coarse and may include spurious regions, the intermediate image in panel (e) preserves the source head pose and shape.
This enables Stage~2 to extract a pixel-level hair mask $M_{\mathrm{seg}}$ in panel (f), which localizes diffusion inpainting to the hair region, reduces unintended changes to non-hair content, and yields a well-aligned final result in panel (g).

For comparison, we apply the same stage-wise mask extraction and visualization procedure to the IP-Adapter baseline.
Using the standard $t_8$-based extraction, IP-Adapter produces diffuse, low-contrast attention in panel (c), and the resulting binarized mask in panel (d) becomes almost entirely active, leading to near-global edits in the intermediate image in panel (e).

Since $t_8$ does not provide a reliable separator token for IP-Adapter, we additionally report results using a mask derived from $t_2$.
Using $t_2$, which consistently emphasizes facial regions in panel (b-1), yields a mask that preserves the face while inpainting most non-facial regions in panel (d), inducing silhouette drift in panel (e) and propagating to a misaligned $M_{\mathrm{seg}}$ in panel (f) with boundary drift in the final output in panel (g).

\section{Implementation Details}
\label{sup:detail}

\subsection{Training Details}
\label{sup:detail:training}
Following \cref{sec:exp:experimental-setup}, we use self-pairs for FFHQ~\cite{karras2019style} and frame pairs from the same identity for CelebV-HQ~\cite{zhu2022celebv}. For CelebV-HQ, we use the official metadata to retrieve the corresponding YouTube videos and retain only clips from which valid training pairs can be constructed. From each valid clip, we sample every 5th frame, estimate head pose for each sampled frame, and compute the yaw angle. Training pairs are then constructed by selecting, for each identity, the frame pair with the largest yaw difference. We further exclude samples without valid hair masks, \ie, cases where BiSeNet~\cite{yu2018bisenet} fails to produce a hair mask. The resulting training set contains 68{,}058 FFHQ images and 9{,}188 CelebV-HQ pairs. For text conditioning during training, we use the fixed generic prompt ``a photo of a person'' for all samples, so that hairstyle information is provided primarily through the reference-image condition rather than through text. Training is conducted for 16{,}000 steps on a single NVIDIA RTX 5090 GPU with batch size 8 and learning rate $1\times10^{-4}$.

\begin{table}[t]
\centering
\caption{Average inference runtime per source--reference pair.}
\label{tab:sup:runtime}
\begin{tabular}{lc}
\toprule
Method & Runtime (s) $\downarrow$ \\
\midrule
HairFastGAN & 0.78 \\
H-Adapter (ours) & 3.05 \\
Stable-Hair & 8.82 \\
HairFusion & 66.28 \\
\bottomrule
\end{tabular}
\end{table}

\subsection{Inference Details}
\label{sup:detail:inference}
In all experiments involving H-Adapter at inference time, we use CLIP-ViT-H-14~\cite{radford2021learning, clip_vit_h_14} as the image encoder.
H-Adapter is initialized from pretrained IP-Adapter-Plus weights~\cite{ye2023ip, ipadapter}, which use patch-based image conditioning.
Although H-Adapter is trained on a Stable Diffusion base model, it can be attached at inference time to compatible variants that share the same base architecture.

\subsubsection{Inpainting-based Hairstyle Transfer}
\label{sup:detail:hairtransfer}
For the inpainting-based hairstyle transfer pipeline, we use Stable Diffusion v1.5 Inpainting~\cite{rombach2022high, sd15_inpainting} as the diffusion backbone.
The hair-removed base images provided to the inpainting model are synthesized using FLUX.2-klein-9B~\cite{bfl2026flux2klein}.
The exact prompt used for synthesizing the hair-removed base images is provided in \cref{sup:sec:flux_prompt}.

\subsubsection{Reference-Guided Text-to-Image Generation}
\label{sup:detail:t2i}
For the reference-guided text-to-image generation results in \cref{sec:method:extensions} of the main paper, we use a standard text-to-image pipeline rather than the inpainting model, with Realistic Vision V4.0 ~\cite{realistic_vision_v4_novae} as the base model and a VAE fine-tuned with an MSE objective~\cite{sd_vae_ft_mse}.
Notably, the H-Adapter trained once with the default SD base model can be directly attached to the variant pipeline, where we apply the per-timestep reference gating described in \cref{sec:method:inference} of the main paper during inference.

\subsubsection{Compatibility with Identity-Preserving Adapters}
\label{sup:detail:faceid}
We use the same text-to-image pipeline, base model, VAE, and H-Adapter reference gating as above.
In this setting, H-Adapter is combined with IP-Adapter FaceID Plus~\cite{ipadapter_faceid} for identity preservation.
We further apply complementary spatial masking: the H-Adapter branch is gated by the hair mask, while the IP-Adapter FaceID Plus branch is gated by the complementary non-hair mask. The effect of this design choice is analyzed in \cref{sup:results:plus}.

\begin{figure}[t]
    \centering
    \includegraphics[width=\linewidth]{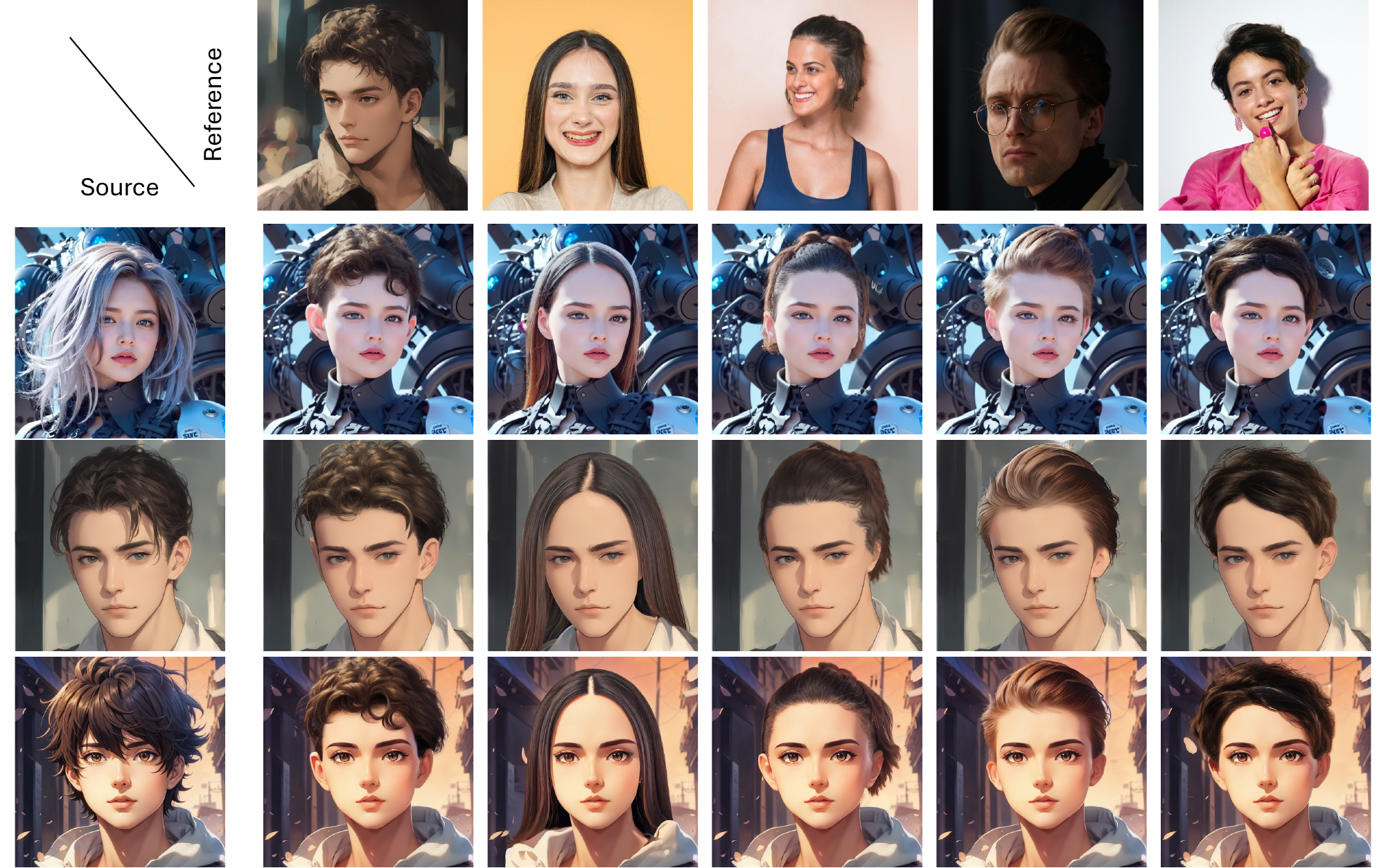}
    \caption{Additional qualitative results on stylized in-the-wild images. The proposed method generalizes well to illustration- and animation-like portraits. Source and license information for the Pexels and Pixabay images used in this figure
is provided in \cref{tab:sup:stock-sources}}
    \label{sup:fig:diverse_domain}
\end{figure}

\begin{figure}[t]
    \centering
    \includegraphics[width=\linewidth]{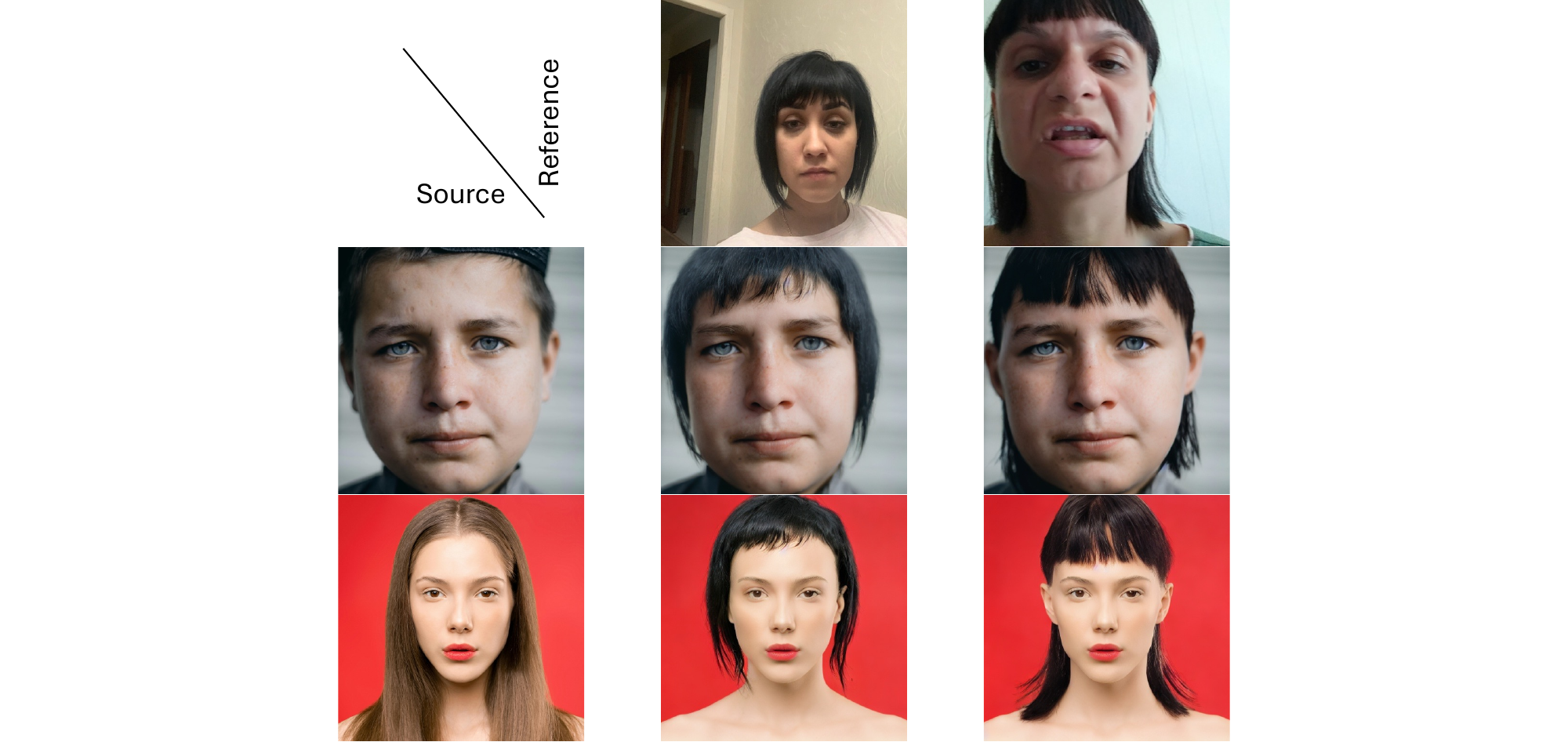}
    \caption{Observed limitation on in-the-wild reference images from EasyPortrait~\cite{kvanchiani2023easyportrait}. In these cases, the generated front hair does not fully preserve the relative placement and facial coverage pattern of the reference. Input images adapted from Pexels stock images under the Pexels License.}
    \label{sup:fig:results:in_the_wild:failure}
\end{figure}

\subsection{Runtime Analysis}
As shown in \cref{tab:sup:runtime}, H-Adapter is substantially faster than the compared diffusion-based hairstyle transfer methods while remaining slower than HairFastGAN, an encoder-based GAN model. The additional inference cost arises from our warm-up and coarse-to-fine inpainting pipeline, which improves localization accuracy and reference fidelity. We do not include earlier optimization-based GAN methods because they require costly per-instance optimization and are therefore not directly comparable in terms of inference latency.

\begin{figure}[t]
    \centering
    \includegraphics[width=\linewidth]{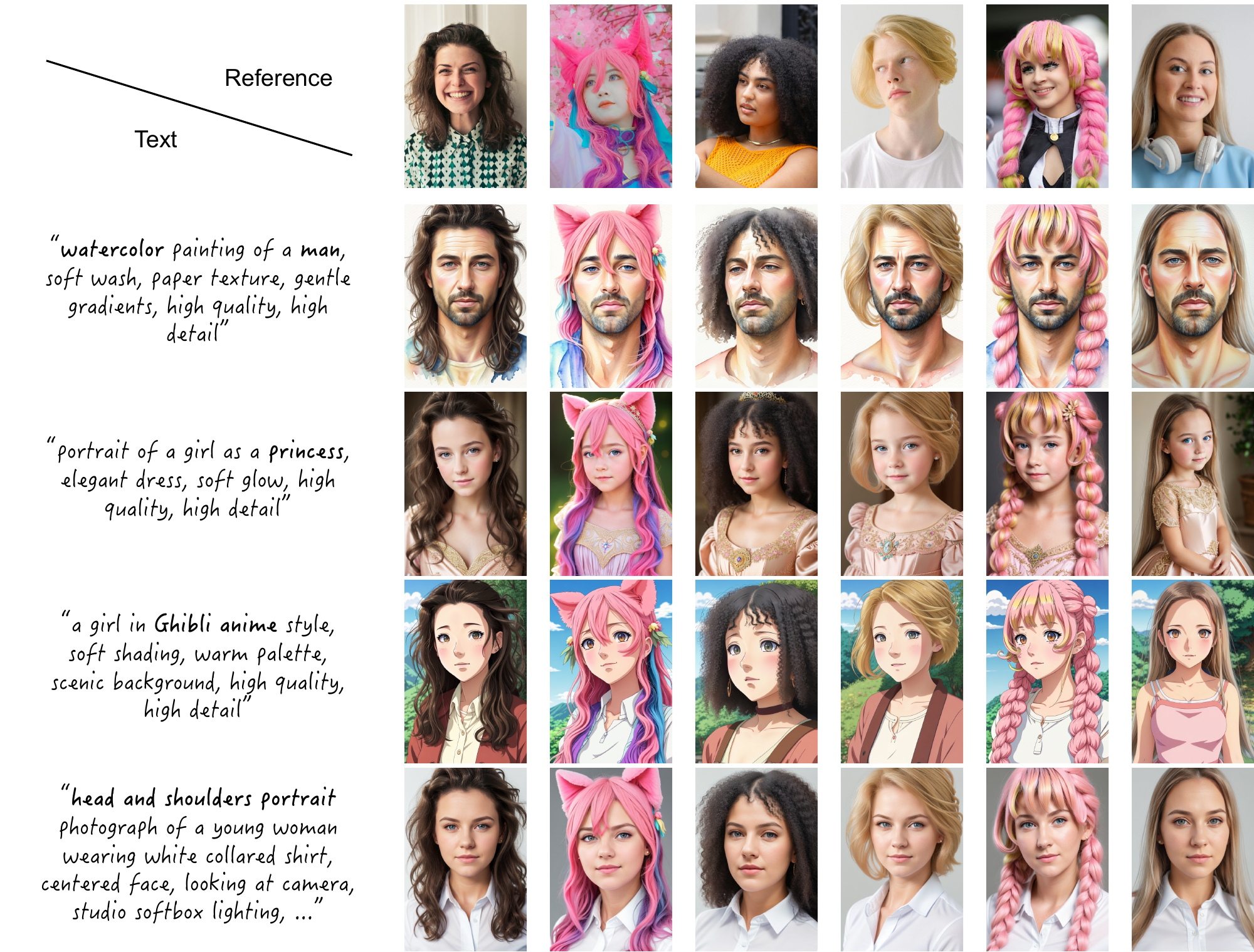}
    \caption{Reference-guided text-to-image generation with H-Adapter. Given diverse text prompts and reference hairstyle images, H-Adapter generates images that reflect both the textual prompt and the key hairstyle attributes of the reference image. All results are shown with the H-Adapter conditioning scale $\lambda=0.6$. Input images adapted from Pexels stock images under the Pexels License.}
    \label{sup:fig:results:t2i}
\end{figure}

\begin{figure}[t]
    \centering
    \includegraphics[width=\linewidth]{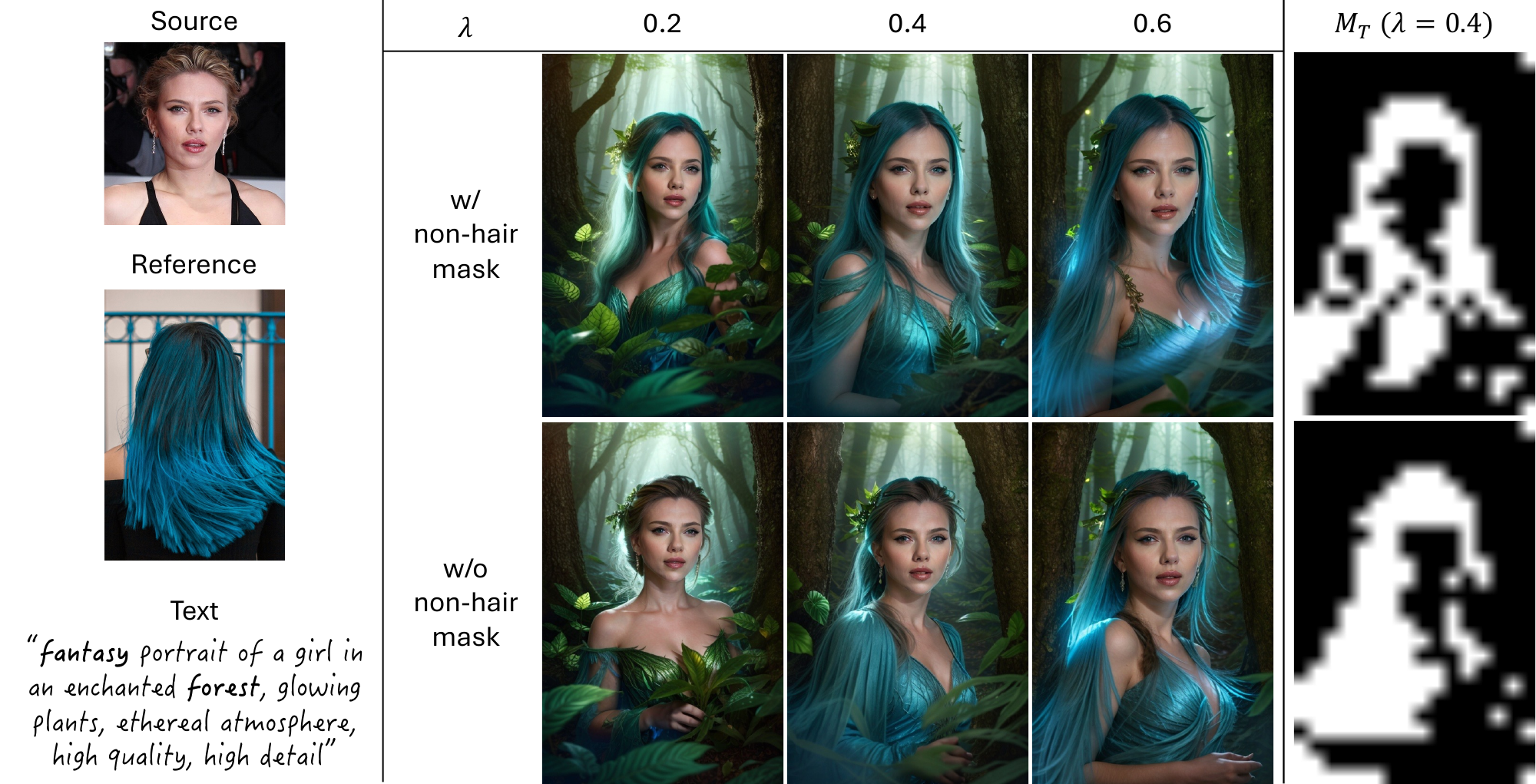}
    \caption{Compatibility of H-Adapter with IP-Adapter FaceID Plus. Results are generated using a text-to-image pipeline with both H-Adapter and IP-Adapter FaceID Plus. The middle panels compare results w/ non-hair mask and w/o non-hair mask on the FaceID Plus branch across different H-Adapter conditioning scales $\lambda$. The rightmost panel visualizes the gating mask $M_T$ extracted at the final denoising step $T$ for the example with $\lambda=0.4$. Input images are adapted from \textcopyright{} JCS, Wikimedia Commons, licensed under CC BY 3.0, and from Pexels stock images under the Pexels License.}
    \label{sup:fig:results:plus}
\end{figure}

\section{Additional Qualitative Results}
\label{sup:results}

\Cref{sup:fig:diverse_domain} additionally presents results not only on real photographs but also on stylized images.
Although H-Adapter is trained exclusively on real-photograph data from FFHQ and CelebV-HQ, it readily extends to these out-of-domain stylized inputs without domain adaptation or additional training, partly due to the broad image-conditioning prior inherited from pretrained IP-Adapter-Plus~\cite{ye2023ip, ipadapter}.
Despite the domain shift, it still produces plausible hair placement while reflecting key attributes of the reference hairstyle.
These results suggest that the proposed approach generalizes to a broader range of visual domains.

\subsubsection{Observed Limitation.}
\label{sup:results:in_the_wild:limit}
Across diverse in-the-wild examples, our method produces plausible hair placement and overall hairstyle transfer under varying poses and appearances.
Nevertheless, we also observe failure cases in \cref{sup:fig:results:in_the_wild:failure}, where the model fails to preserve the reference-specific spatial relationship between hair and the underlying facial structure. In particular, although the generated hair may appear plausible on its own, it may not accurately reflect how the bangs extend relative to the forehead or eyebrow region in the reference. These observations suggest that the remaining limitation lies in modeling the fine-grained spatial relationship between hairstyle and facial geometry.

\subsection{Reference-Guided Text-to-Image Generation}
\label{sup:results:t2i}
We further apply H-Adapter to a standard text-to-image pipeline to generate images conditioned on a text prompt and a reference hairstyle. As shown in \cref{sup:fig:results:t2i}, the generated images reflect both the textual prompt and the key hairstyle attributes of the reference image. These results indicate that H-Adapter can support reference-guided hairstyle-aware generation across diverse textual contexts.

\subsection{Compatibility with Identity-Preserving Adapters}
\label{sup:results:plus}
We analyze the effect of the complementary spatial masking described in \cref{sup:detail:faceid}. In \cref{eq:method:ref-gated-attn} of the main paper, the H-Adapter branch is gated by the hair mask $M_t$. When combining H-Adapter with IP-Adapter FaceID Plus, we extend the formulation by additionally applying the complementary mask $1-M_t$ to the FaceID Plus branch:
\begin{equation}
    \label{eq:sup:method:ref-gated-attn-faceid}
    \begin{aligned}
    Z
    &= \mathrm{softmax}\!\left(\frac{QK^{\top}}{\sqrt{d}}\right)V \\
    &\quad + \lambda \Bigl(M_t \odot \mathrm{softmax}\!\left(\frac{Q{K'}^{\top}}{\sqrt{d}}\right)V'\Bigr) \\
    &\quad + \lambda_{\mathrm{face}} \Bigl((1-M_t) \odot \mathrm{softmax}\!\left(\frac{Q{K''}^{\top}}{\sqrt{d}}\right)V''\Bigr),
    \end{aligned}
\end{equation}
where $(K, V)$ denote the key/value pairs from the standard cross-attention context, $(K', V')$ denote those from the H-Adapter branch, and $(K'', V'')$ denote those from the IP-Adapter FaceID Plus branch. Here, $\lambda$ controls the strength of the hairstyle condition from H-Adapter, while $\lambda_{\text{face}}$ controls the strength of the identity-preserving condition from the FaceID Plus branch. As shown in \cref{sup:fig:results:plus}, this complementary masking leads to better reflection of the reference hairstyle across different values of the H-Adapter conditioning scale $\lambda$. In contrast, without applying $1-M_t$ to the identity-preserving branch, the FaceID Plus condition more strongly interferes with the hairstyle condition, resulting in weaker reflection of the reference hairstyle.

For visualization, we also show the gating mask $M_T$ extracted for the H-Adapter branch at the final denoising step $T$ for the example with $\lambda=0.4$, while fixing the FaceID Plus conditioning scale to $\lambda_{\mathrm{face}}=0.6$. The mask illustrates the spatial support of the H-Adapter branch, primarily covering the hair region. All results in \cref{fig:exp:personalization} of the main paper are generated with this complementary masking applied to the FaceID Plus branch.


\begin{table}[t]
\centering
\caption{Summary of the pointwise VLM-as-a-judge prompt schema. Full verbatim prompts are provided in the
supplementary prompt files.}
\label{tab:vlm_prompt_schema}
\scriptsize
\setlength{\tabcolsep}{3pt}
\begin{tabular}{p{0.08\linewidth}p{0.16\linewidth}p{0.28\linewidth}p{0.20\linewidth}p{0.22\linewidth}}
\toprule
Axis & Images & Evaluate & Ignore & Allowed tags \\
\midrule
HFS &
Reference, Result &
Hair color, tone, highlights, texture, length, silhouette, volume, shape, bangs, parting, forehead exposure,
hairline consistency &
Face identity, skin, expression, background, clothing, accessories, global beauty unless hair visibility is
affected &
\texttt{silhouette\_mismatch},
\texttt{bangs\_part\_mismatch},
\texttt{texture\_mismatch},
\texttt{color\_mismatch},
\texttt{hairline\_mismatch},
\texttt{partial\_transfer},
\texttt{style\_mismatch}
\\
\midrule
NPS &
Source, Result &
Facial identity, facial geometry, skin tone, local lighting, expression, pose, background, clothing,
accessories, non-hair distortions that affect identity or scene preservation &
Hairstyle quality and hairstyle similarity to the reference &
\texttt{identity\_drift},
\texttt{facial\_feature\_change},
\texttt{skin\_tone\_shift},
\texttt{lighting\_shift},
\texttt{pose\_expression\_change},
\texttt{background\_change},
\texttt{clothing\_accessory\_change},
\texttt{artifact\_or\_distortion}
\\
\midrule
AQS &
Result only &
Hair boundary blending, hairline or contour artifacts, unnatural face or skin texture, visible seams,
compositing traces, structural distortion, blur, noise, compression, repetition artifacts &
Hair-reference similarity, source identity preservation, personal aesthetic preference &
\texttt{boundary\_blending\_artifact},
\texttt{hairline\_contour\_artifact},
\texttt{unnatural\_skin\_texture},
\texttt{visible\_patch\_or\_seam},
\texttt{structural\_distortion},
\texttt{blur\_or\_focus\_artifact},
\texttt{noise\_or\_compression\_artifact},
\texttt{repetition\_or\_tiling\_artifact}
\\
\bottomrule
\end{tabular}
\end{table}

\begin{table}[t]
\centering
\caption{Common 1--5 score anchors used by the pointwise prompts. The axis-specific wording instantiates
these anchors for hair fidelity, non-hair preservation, or artifact quality.}
\label{tab:vlm_score_anchors}
\scriptsize
\begin{tabular}{cl}
\toprule
Score & Anchor meaning \\
\midrule
5 & Nearly perfect result with no meaningful axis-specific error. \\
4 & Strong result with only minor axis-specific inconsistency. \\
3 & Partial success with clear but moderate axis-specific errors. \\
2 & Weak result with multiple major axis-specific errors. \\
1 & Failure case: largely unrelated, poorly preserved, or severely degraded result. \\
\bottomrule
\end{tabular}
\end{table}

\begin{table}[t]
\centering
\caption{Pointwise prompt variants. Each axis uses all three variants, yielding nine pointwise templates in
total.}
\label{tab:vlm_prompt_variants}
\scriptsize

\begin{tabular}{cll}
\toprule
Variant & Prompt framing & Purpose \\
\midrule
A & Axis goal with explicit evaluation scope and full rubric & Standard checklist-based scoring. \\
B & Alternative checklist wording with the same score anchors & Reduces sensitivity to a single phrasing. \\
C & Error-tag-first procedure followed by severity-to-score mapping & Encourages evidence-grounded scoring. \\
\bottomrule
\end{tabular}
\end{table}

\section{VLM-as-a-Judge}
\label{sup:vlm}

\subsection{Evaluation Protocol}
\label{sup:vlm_prompt}
 
We evaluate each generated result along three independent axes:
\textsc{Hair Fidelity Score (HFS)},
\textsc{Non-hair Preservation Score (NPS)}, and
\textsc{Artifact Quality Score (AQS)}.
To reduce inter-axis interference~\cite{ku2024viescore}, we score each axis independently using an axis-specific set of pointwise prompts rather than a joint multi-criteria prompt.

For each axis, we use three prompt variants that share the same evaluation objective, output schema, and 1--5 Likert anchors~\cite{lee2024prometheus}, but differ in both surface wording and evaluation framing. In particular, Template C asks the judge to identify diagnostic error tags before mapping severity to a final score. This design reduces sensitivity to any single prompt formulation~\cite{slyman2025calibrating}. Each triplet is evaluated independently with all three variants, and the scores are averaged to obtain the \emph{per-triplet template-mean score} for that axis. Method-level results are then computed by averaging these template-mean scores over the 200-triplet evaluation set.
 
We run the same evaluation pipeline with three VLM judges: GPT-4o~\cite{hurst2024gpt4o}, GPT-5.2~\cite{openai2025gpt5, openai2025gpt52}, and Gemini-2.5-Flash~\cite{comanici2025gemini25}. All three judges use versioned API snapshots for reproducibility. To isolate judge behavior from prompt wording, all judges receive identical prompts and image orderings.

\subsection{Pointwise Prompt Schema and Templates}
\label{sup:sec:exact-prompts:vlm}

All pointwise prompts follow a shared schema. \Cref{tab:vlm_prompt_schema} summarizes the axis-specific image inputs, evaluation scope, ignored factors, and diagnostic tags; \Cref{tab:vlm_score_anchors} reports the common 1--5 score anchors; and \cref{tab:vlm_prompt_variants} describes the three prompt variants used for each axis. The full verbatim prompts used in the experiments are provided as supplementary machine-readable text files.

Each pointwise prompt contains: (i) a role instruction, (ii) the target evaluation axis, (iii) the fixed image order, (iv) the axis-specific evaluation scope, (v) factors to ignore, (vi) a 1--5 Likert rubric, (vii) allowed diagnostic tags, (viii) an uncertainty rule, and (ix) a JSON-only output constraint. Template~A uses an explicit attribute checklist, Template~B uses an alternative checklist phrasing, and Template~C uses an error-tag-first procedure that maps error severity to a score.

All pointwise prompts require the judge to return a JSON object with the same fields:
\begin{Verbatim}[breaklines=true, breakanywhere=true, fontsize=\scriptsize]
{
"axis": "<HFS|NPS|AQS>",
"rationale": ["<short visual evidence 1>", "<short visual evidence 2>"],
"tags": ["<diagnostic tag 1>", "<diagnostic tag 2>"],
"is_uncertain": <true|false>,
"score": <integer 1..5>
}
\end{Verbatim}
The rationale field is restricted to two to four short visual evidence statements, the score field must be an integer from 1 to~5, and no text outside the JSON object is allowed. If the relevant visual evidence is insufficient due to occlusion, blur, extreme pose, or severe cropping, the judge is instructed to set \texttt{is\_uncertain=true}.

\begin{table}[t]
\centering
\caption{
Consistency validation across judge models. The same 80 evaluation queries are repeated three times per judge (240 queries total), where each query is a unique (source, reference, generated output, method, axis) tuple. We report Krippendorff's $\alpha$ (ordinal) and the mean per-query score standard deviation ($\sigma$) per axis.
}

\label{tab:consistency_audit_models}
\begin{tabular}{l>{\centering\arraybackslash}p{0.1\linewidth}>{\centering\arraybackslash}p{0.22\linewidth}>{\centering\arraybackslash}p{0.12\linewidth}}
\toprule
Judge Model & Axis & Krippendorff's $\alpha\uparrow$ & $\sigma\downarrow$ \\
\midrule
\multirow{3}{*}{GPT-4o}           & HFS & 0.985 & 0.020 \\
                                  & NPS & 0.902 & 0.067 \\
                                  & AQS & 0.991 & 0.023 \\
\midrule
\multirow{3}{*}{GPT-5.2}          & HFS & 0.884 & 0.159 \\
                                  & NPS & 0.364 & 0.200 \\
                                  & AQS & 0.804 & 0.208 \\
\midrule
\multirow{3}{*}{Gemini-2.5-Flash} & HFS & 0.902 & 0.093 \\
                                  & NPS & 0.886 & 0.162 \\
                                  & AQS & 0.952 & 0.048 \\
\bottomrule
\end{tabular}
\end{table}

\subsection{Reliability and Validation Protocols}
\label{sup:vlm_reliability}

We validate the VLM-as-a-judge pipeline with three complementary procedures: (i) a test--retest consistency audit, (ii) a bidirectional pairwise preference audit, and (iii) cross-judge execution with the three VLM judges defined above.

\subsubsection{Consistency Validation.}

This audit measures how reproducible each judge is when the same query is scored repeatedly. We sample 80 scoring queries across the three axes, each corresponding to a unique (source, reference, generated output, method, axis) tuple from the main evaluation set, and re-evaluate every query three times. All repeats use a single fixed template (Template~A), so the audit estimates stochastic test--retest reliability for a fixed prompt. Treating the repeats as raters, we report ordinal Krippendorff's $\alpha$ and the mean per-query $\sigma$ (\cref{tab:consistency_audit_models}).

GPT-4o is the most reproducible judge ($\alpha \ge 0.90$, $\sigma \le 0.07$ on all axes), supporting its use as the representative judge in the main paper (\cref{tab:mean_scores_combined}). Gemini-2.5-Flash is likewise stable ($\alpha \ge 0.886$). GPT-5.2 remains acceptable on HFS ($\alpha = 0.884$) and AQS ($\alpha = 0.804$) but drops on NPS ($\alpha = 0.364$, $\sigma = 0.200$), reflecting run-to-run variability in fine-grained identity judgments. Because the NPS audit contains only 26 unique items (78 evaluations across three repeats), this estimate should be treated as indicative rather than precise; we therefore treat GPT-5.2 NPS results---pointwise (\cref{tab:scores_gpt52}) and pairwise (\cref{tab:pairwise_winrate_all})---as low-reliability throughout. Because the main scores average three fixed prompt variants, the single-template $\sigma$ is a conservative proxy for the residual run-to-run noise of the template-averaged scores: averaging the stochastic component measured here should not increase its variance and typically reduces it.

\begin{table}[t]
\centering
\caption{VLM-as-a-judge scores (GPT-5.2) with 95\,\% bootstrap
         percentile CIs ($n_{\text{boot}}=1{,}000$ over per-triplet
         template-mean scores).}
\label{tab:scores_gpt52}
\begin{tabular}{lccc}
\toprule
Method & HFS$\uparrow$ & NPS$\uparrow$ & AQS$\uparrow$ \\
\midrule
Ours            & \textbf{3.72}~{\scriptsize[3.62, 3.81]} & \textbf{4.16}~{\scriptsize[4.08, 4.23]} & \textbf{3.48}~{\scriptsize[3.39, 3.56]} \\
\midrule
HairFusion      & 2.95~{\scriptsize[2.82, 3.08]} & 3.56~{\scriptsize[3.48, 3.65]} & 3.02~{\scriptsize[2.92, 3.13]} \\
Stable-Hair     & 3.58~{\scriptsize[3.46, 3.70]} & 3.58~{\scriptsize[3.49, 3.68]} & 2.93~{\scriptsize[2.83, 3.03]} \\
HairFastGAN     & 3.30~{\scriptsize[3.17, 3.42]} & 4.10~{\scriptsize[4.04, 4.17]} & 3.40~{\scriptsize[3.31, 3.50]} \\
HairCLIPv2      & 2.37~{\scriptsize[2.22, 2.51]} & 4.04~{\scriptsize[3.96, 4.11]} & 3.42~{\scriptsize[3.32, 3.52]} \\
Style-Your-Hair & 3.33~{\scriptsize[3.21, 3.45]} & 4.00~{\scriptsize[3.93, 4.07]} & 3.38~{\scriptsize[3.28, 3.47]} \\
\bottomrule
\end{tabular}
\end{table}
 
\begin{table}[t]
\centering
\caption{VLM-as-a-judge scores (Gemini-2.5-Flash) with 95\,\% bootstrap
         percentile CIs ($n_{\text{boot}}=1{,}000$ over per-triplet
         template-mean scores).}
\label{tab:scores_gemini25flash}
\begin{tabular}{lccc}
\toprule
Method & HFS$\uparrow$ & NPS$\uparrow$ & AQS$\uparrow$ \\
\midrule
Ours            & \textbf{2.68}~{\scriptsize[2.54, 2.81]} & \textbf{3.11}~{\scriptsize[2.98, 3.24]} & \textbf{2.74}~{\scriptsize[2.62, 2.86]} \\
\midrule
HairFusion      & 2.08~{\scriptsize[1.98, 2.20]} & 2.39~{\scriptsize[2.29, 2.50]} & 2.08~{\scriptsize[1.99, 2.18]} \\
Stable-Hair     & 2.28~{\scriptsize[2.14, 2.42]} & 2.25~{\scriptsize[2.14, 2.36]} & 1.91~{\scriptsize[1.82, 2.00]} \\
HairFastGAN     & 2.41~{\scriptsize[2.30, 2.52]} & 2.79~{\scriptsize[2.69, 2.90]} & 2.38~{\scriptsize[2.29, 2.48]} \\
HairCLIPv2      & 1.80~{\scriptsize[1.68, 1.91]} & 2.87~{\scriptsize[2.75, 2.99]} & 2.45~{\scriptsize[2.33, 2.56]} \\
Style-Your-Hair & 2.40~{\scriptsize[2.26, 2.54]} & 2.76~{\scriptsize[2.64, 2.86]} & 2.51~{\scriptsize[2.40, 2.60]} \\
\bottomrule
\end{tabular}
\end{table}
 
\subsubsection{Pairwise Preference Protocol.}
As a complementary check on the pointwise rankings reported in the main paper, we conduct a pairwise preference analysis on a separate subset of 40~triplets. For each triplet, the judge is presented with two outputs---ours and one baseline---and asked to select a winner (A, B, or Tie) on each axis. To control for position bias, we use a bidirectional protocol: the same pair is evaluated twice with the presentation order reversed (AB and BA). If the two orderings yield a consistent winner, the verdict is accepted; otherwise the comparison is recorded as a Tie.

Win rates in \cref{tab:pairwise_winrate_all} are computed as $\text{wins}_{\text{ours}} / n$, where ties count as neither a win nor a loss.
\footnote{%
  An alternative convention assigns half-credit to ties
  ($\text{adjusted} = (\text{wins} + 0.5 \times \text{ties}) / n$).
  We report the strict variant for transparency; the adjusted values
  lead to the same ordinal conclusions.} 

\subsection{Pointwise Results Across Judges}
\label{sup:vlm_pointwise_results}

\Cref{tab:scores_gpt52,tab:scores_gemini25flash} report the full pointwise scores for GPT-5.2 and Gemini-2.5-Flash, respectively, complementing the GPT-4o results in \cref{tab:mean_scores_combined} of the main paper.

\subsubsection{Ranking Consistency and Calibration.}
Across all three judges, the top of the ranking is stable: ours attains the highest mean on every axis, and no baseline matches it on all three axes simultaneously. Two per-baseline trends are likewise judge-robust: HairCLIPv2~\cite{wei2023hairclipv2} ranks last on HFS while remaining competitive (second or third) on both NPS and AQS, and Stable-Hair~\cite{zhang2025stable} ranks last on AQS.

The relative ordering among the remaining mid-ranked baselines is judge-dependent and should not be over-interpreted. For example, Stable-Hair is second on HFS under GPT-4o and GPT-5.2 but fourth under Gemini-2.5-Flash; its NPS is the lowest under GPT-4o and Gemini-2.5-Flash but marginally above HairFusion~\cite{chung2025preserve} under GPT-5.2 (3.58 vs.\ 3.56), for which GPT-5.2's NPS reliability is low (\cref{tab:consistency_audit_models}). Similarly, Style-Your-Hair~\cite{kim2022style} is the most balanced competitor---second or third on AQS under GPT-4o and Gemini-2.5-Flash and competitive on NPS---but drops to fourth on AQS under GPT-5.2. These small reorderings partly reflect cross-judge calibration differences; the consistent conclusion is that ours is the only method that remains top-tier on all three axes regardless of judge.

Gemini-2.5-Flash assigns systematically lower absolute scores than the other two judges (\eg, the highest HFS mean is 2.68 vs.\ 3.11 for GPT-4o and 3.72 for GPT-5.2), reflecting a stricter interpretation of the rubric anchors rather than a disagreement in the main ordinal conclusions. GPT-5.2 tends to assign higher absolute scores in some settings, but the shift is axis- and method-dependent. These calibration differences motivate reporting ordinal rankings and within-judge comparisons rather than comparing raw scores across judges.

\begin{table}[t]
\centering
\caption{Pairwise win rates of \textit{ours} vs.\ each baseline on the
         40-triplet validation subset (bidirectional protocol, 
         \(\text{win rate} = \text{wins}_{\text{ours}} / n\);
         ties counted as neither win nor loss).
         Results are shown for three VLM judges.}
\label{tab:pairwise_winrate_all}
\begin{tabular}{@{} l *{9}{>{\centering\arraybackslash}p{0.083\linewidth}} @{}}
\toprule
& \multicolumn{3}{c}{\textbf{GPT-4o}} 
& \multicolumn{3}{c}{\textbf{Gemini-2.5-Flash}} 
& \multicolumn{3}{c}{\textbf{GPT-5.2}} \\
\cmidrule(lr){2-4} \cmidrule(lr){5-7} \cmidrule(lr){8-10}
Method & HFS & NPS & AQS
       & HFS & NPS & AQS
       & HFS & NPS & AQS \\
\midrule
HairFusion      & 0.71 & 0.93 & 0.80 & 0.70 & 0.91 & 0.74 & 0.84 & 0.86 & 0.79 \\
Stable-Hair     & 0.64 & 0.91 & 0.85 & 0.61 & 0.85 & 0.88 & 0.50 & 0.80 & 0.81 \\
HairFastGAN     & 0.63 & 0.73 & 0.68 & 0.63 & 0.61 & 0.59 & 0.66 & 0.63 & 0.55 \\
HairCLIPv2      & 0.88 & 0.56 & 0.54 & 0.78 & 0.60 & 0.58 & 0.91 & 0.63 & 0.49 \\
Style-Your-Hair & 0.63 & 0.61 & 0.48 & 0.55 & 0.63 & 0.50 & 0.65 & 0.69 & 0.56 \\
\bottomrule
\end{tabular}
\end{table}

\subsection{Pairwise Results Across Judges}
\label{sup:vlm_pairwise_results}

\subsubsection{Overall Pairwise Results.}
The pairwise win rates in \cref{tab:pairwise_winrate_all} provide an independent validation of the pointwise rankings. Across all three judges, ours achieves win rates above 0.50 on nearly every axis--baseline combination, with large margins on HFS (\eg, 0.88 vs.\ HairCLIPv2 with GPT-4o) and NPS (\eg, 0.93 vs.\ HairFusion with GPT-4o).

The per-baseline trade-offs identified in the pointwise analysis are mirrored in the pairwise results. Against HairCLIPv2, ours achieves dominant HFS win rates ($\geq 0.78$), but the margin narrows on NPS and AQS (0.49--0.63), reflecting HairCLIPv2's conservative editing strategy. Against Stable-Hair, the pattern reverses: HFS win rates are among the most modest (0.50--0.64), while ours dominates on NPS (0.80--0.91) and AQS (0.81--0.88). The closest overall competitor remains Style-Your-Hair, whose AQS win rates reach near-parity at 0.48 (GPT-4o) and 0.50 (Gemini-2.5-Flash), consistent with the overlapping confidence intervals reported in the main paper; however, ours maintains a clear HFS advantage (0.55--0.65).

With the reliability qualification in \cref{sup:vlm_reliability}, the pairwise trends remain aligned with the other two judges.

\medskip\noindent
Overall, the pairwise results provide an independent validation of the pointwise analysis in the main paper. Across judges, the directional preferences are broadly consistent with the pointwise score differences, supporting the same qualitative conclusions.

\begin{table}[t]
\centering
\caption{Comparison with FLUX.2 on the pose-different subset.}
\label{sup:tab:quant:flux2}
\resizebox{0.72\linewidth}{!}{
\begin{tabular}{lccccc}
\toprule
Method & FID $\downarrow$ & FID$_{\mathrm{CLIP}}$ $\downarrow$ & SSIM $\uparrow$ & PSNR $\uparrow$ & CLIP-I $\uparrow$ \\
\midrule
Ours   & \textbf{12.47} & \textbf{3.98} & 0.831 & 23.06 & \textbf{0.659} \\
FLUX.2 & 12.66 & 5.18 & \textbf{0.904} & \textbf{25.35} & 0.643 \\
\bottomrule
\end{tabular}
}
\end{table}


\begin{table}[t]
\centering
\caption{Quantitative comparison on the pose-agnostic subset.}
\label{sup:tab:pa}
\begin{tabular}{lccccc}
\toprule
Method & FID $\downarrow$ & FID$_{\mathrm{CLIP}}$ $\downarrow$ & SSIM $\uparrow$ & PSNR $\uparrow$ & CLIP-I $\uparrow$ \\
\midrule
Ours                       & \textbf{11.56} & \textbf{3.97} & \textbf{0.833} & 23.25 & \textbf{0.669} \\
\midrule
IP-Adapter ($t_8$)          & 12.94 & 6.98 & 0.815 & 22.33 & 0.656 \\
IP-Adapter ($t_2$)          & 11.88 & 4.14 & 0.829 & 23.04 & 0.662 \\
\midrule
HairFusion                 & 28.92 & 9.07 & 0.748 & 16.87 & 0.632 \\
Stable-Hair                & 25.98 & 8.16 & 0.797 & 22.32 & 0.652 \\
HairFastGAN                & 12.30 & 4.44 & 0.815 & \textbf{24.46} & 0.661 \\
HairCLIPv2                 & 13.30 & 6.65 & 0.831 & 24.18 & 0.635 \\
Style-Your-Hair            & 15.53 & 7.58 & 0.821 & 23.19 & 0.661 \\
\bottomrule
\end{tabular}
\end{table}

\section{Additional Quantitative Analysis}
\label{sup:quant}

\subsection{Comparison with a General-Purpose Image Editor}
\label{sup:quant:flux2}

We additionally compare H-Adapter with FLUX.2~\cite{bfl2026flux2klein}, a recent general-purpose image editing model, on the pose-different subset. 
Since general-purpose editors and task-specific hairstyle-transfer methods differ in their intended use and evaluation focus, we report this comparison separately from the main quantitative table.

As shown in \cref{sup:tab:quant:flux2}, FLUX.2 achieves higher SSIM and PSNR, reflecting its stronger tendency to preserve the source appearance. However, as illustrated in the qualitative comparisons in \cref{fig:qualitative-comparison} in the main paper, such pixel-level similarity does not necessarily translate into faithful reference-hairstyle transfer. FLUX.2 often produces visually plausible edits that remain closer to the source hairstyle, while failing to reproduce reference-specific hairstyle attributes under pose mismatch. In contrast, H-Adapter achieves better FID, FID$_{\mathrm{CLIP}}$, and CLIP-I, indicating higher reference-hairstyle fidelity and perceptual quality while maintaining competitive preservation of non-hair content.

\begin{table}[t]
\centering
\caption{Quantitative comparison under the controlled evaluation with hair-removed source images (FLUX.2). For comparison methods, values in parentheses denote differences from the corresponding original-setting results on the same reduced subset.}
\label{sup:tab:flux_comparison}
\setlength{\tabcolsep}{3.5pt}
\begin{tabular}{lccccc}
\toprule
\multirow{2}{*}{Method} & \multicolumn{5}{c}{Pose-Agnostic ($n=2924$)} \\
\cmidrule(lr){2-6}
 & FID$\downarrow$ & FID$_{\mathrm{CLIP}}\downarrow$ & SSIM$\uparrow$ & PSNR$\uparrow$ & CLIP-I$\uparrow$ \\
\midrule
Ours                    & \textbf{11.59} & \textbf{3.97} & \textbf{0.833} & \textbf{23.24} & \textbf{0.669} \\
IP-Adapter ($t_8$)      & 12.96 & 6.96 & 0.815 & 22.32 & 0.656 \\
IP-Adapter ($t_2$)      & 11.91 & 4.14 & 0.829 & 23.04 & 0.663 \\
HairFusion              & 26.04 {\scriptsize ($-2.99$)} & 9.78 {\scriptsize ($+0.71$)} & 0.725 {\scriptsize ($-0.023$)} & 16.12 {\scriptsize ($-0.74$)} & 0.631 {\scriptsize ($-0.002$)} \\
Stable-Hair             & 17.45 {\scriptsize ($-8.69$)} & 5.99 {\scriptsize ($-2.18$)} & \textbf{0.833} {\scriptsize ($+0.037$)} & 23.15 {\scriptsize ($+0.84$)} & 0.655 {\scriptsize ($+0.003$)} \\
HairFastGAN             & 13.34 {\scriptsize ($+1.03$)} & 5.89 {\scriptsize ($+1.45$)} & 0.794 {\scriptsize ($-0.021$)} & 21.53 {\scriptsize ($-2.93$)} & 0.660 {\scriptsize ($-0.001$)} \\
Style-Your-Hair         & 15.37 {\scriptsize ($-0.19$)} & 8.70 {\scriptsize ($+1.13$)} & 0.811 {\scriptsize ($-0.010$)} & 22.36 {\scriptsize ($-0.83$)} & 0.657 {\scriptsize ($-0.005$)} \\
HairCLIPv2              & 14.56 {\scriptsize ($+1.23$)} & 8.82 {\scriptsize ($+2.18$)} & 0.824 {\scriptsize ($-0.007$)} & 22.99 {\scriptsize ($-1.19$)} & 0.632 {\scriptsize ($-0.003$)} \\
\midrule
\multirow{2}{*}{Method} & \multicolumn{5}{c}{Pose-Different ($n=2876$)} \\
\cmidrule(lr){2-6}
 & FID$\downarrow$ & FID$_{\mathrm{CLIP}}\downarrow$ & SSIM$\uparrow$ & PSNR$\uparrow$ & CLIP-I$\uparrow$ \\
\midrule
Ours                    & \textbf{12.48} & \textbf{3.98} & 0.831 & 23.07 & \textbf{0.659} \\
IP-Adapter ($t_8$)      & 15.29 & 8.82 & 0.803 & 21.56 & 0.640 \\
IP-Adapter ($t_2$)      & 12.55 & 4.26 & 0.825 & 22.70 & 0.652 \\
HairFusion              & 26.35 {\scriptsize ($-1.76$)} & 9.43 {\scriptsize ($+0.63$)} & 0.731 {\scriptsize ($-0.025$)} & 16.50 {\scriptsize ($-0.76$)} & 0.624 {\scriptsize ($-0.002$)} \\
Stable-Hair             & 18.32 {\scriptsize ($-7.56$)} & 6.66 {\scriptsize ($-2.05$)} & \textbf{0.835} {\scriptsize ($+0.037$)} & \textbf{23.19} {\scriptsize ($+0.80$)} & 0.641 {\scriptsize ($+0.001$)} \\
HairFastGAN             & 13.73 {\scriptsize ($+0.92$)} & 6.04 {\scriptsize ($+1.50$)} & 0.795 {\scriptsize ($-0.022$)} & 21.45 {\scriptsize ($-2.96$)} & 0.649 {\scriptsize ($-0.001$)} \\
Style-Your-Hair         & 16.08 {\scriptsize ($+0.09$)} & 10.25 {\scriptsize ($+1.74$)} & 0.807 {\scriptsize ($-0.009$)} & 22.03 {\scriptsize ($-0.78$)} & 0.641 {\scriptsize ($-0.009$)} \\
HairCLIPv2              & 15.43 {\scriptsize ($+1.97$)} & 10.37 {\scriptsize ($+2.48$)} & 0.819 {\scriptsize ($-0.005$)} & 22.63 {\scriptsize ($-1.01$)} & 0.619 {\scriptsize ($-0.004$)} \\
\bottomrule
\end{tabular}
\end{table}

\subsection{Quantitative Results on the Pose-Agnostic Subset}
\label{sup:quant:pa}

We report quantitative results on the pose-agnostic subset in \cref{sup:tab:pa}. 
Unlike the pose-different subset used in the main paper, this subset is constructed without explicitly enforcing a large head-pose gap between the source and reference, and thus reflects a more general hairstyle-transfer setting.

As shown in \cref{sup:tab:pa}, our method achieves the best overall performance, obtaining the best results on FID, FID$_{\mathrm{CLIP}}$, SSIM, and CLIP-I, while remaining competitive in PSNR.
These results indicate that our method generates visually higher-quality images, more faithfully reflects the reference hairstyle, and better preserves non-hair structure.

Compared with the pose-different subset in \cref{tab:pd} in the main paper, most methods perform slightly better on the pose-agnostic subset, which is expected because pose mismatch is less severe.
Nevertheless, the overall trend remains largely unchanged, and our method continues to perform best overall.

\begin{table*}[t]
\centering
\caption{Quantitative comparison for different training data compositions.}
\label{tab:train_data_comp}
\resizebox{\textwidth}{!}{
\begin{tabular}{lccccc|ccccc}
\toprule
& \multicolumn{5}{c|}{Pose-agnostic subset} & \multicolumn{5}{c}{Pose-different subset} \\
\cmidrule(r){2-6} \cmidrule(l){7-11}
Method & FID $\downarrow$ & FID$_{\mathrm{CLIP}}$ $\downarrow$ & SSIM $\uparrow$ & PSNR $\uparrow$ & CLIP-I $\uparrow$
& FID $\downarrow$ & FID$_{\mathrm{CLIP}}$ $\downarrow$ & SSIM $\uparrow$ & PSNR $\uparrow$ & CLIP-I $\uparrow$ \\
\midrule
Ours w/o CelebV-HQ & \textbf{10.97} & 4.25 & 0.763 & 18.35 & 0.668
                    & \textbf{11.79} & 4.22 & 0.761 & 18.35 & 0.657 \\
Ours               & 11.42 & \textbf{3.98} & \textbf{0.833} & \textbf{23.25} & 0.668
                    & 12.34 & \textbf{4.00} & \textbf{0.830} & \textbf{23.05} & 0.657 \\
\bottomrule
\end{tabular}}
\end{table*}

\subsection{Controlled Evaluation with Hair-Removed Source Images}
\label{sup:quant:bald}

We additionally conduct a controlled quantitative evaluation to examine whether the performance gain of H-Adapter can be explained solely by the use of FLUX-based hair-removed source images.
Since our pipeline performs hairstyle transfer on the hair-removed base image $\tilde{x}_s$, this input may provide a favorable condition for preserving non-hair regions.
To isolate this factor, we evaluate all methods under a shared input setting where the original source image is replaced with the same hair-removed base image used by our method.

For HairFusion~\cite{chung2025preserve}, the pipeline includes a preprocessing stage that extracts a source hair mask and constructs a hair-agnostic image prior to generation. We therefore reuse the preprocessing data produced by the original pipeline and replace only the source image with the corresponding hair-removed base image, while keeping all other inputs unchanged.
In this setting, additional failures arise in HairFusion during landmark extraction. We therefore exclude these samples from evaluation, removing an additional 14 samples from the pose-agnostic subset and 23 samples from the pose-different subset relative to the set evaluated in \cref{sup:tab:pa} and \cref{tab:pd}. As a result, this controlled evaluation is conducted on 2{,}924 and 2{,}876 samples for the pose-agnostic and pose-different subsets, respectively. To enable a fair comparison and to make the shift induced by the controlled input setting explicit, we also report, for each comparison method, the difference from its original-setting result on the same reduced subset in parentheses in \cref{sup:tab:flux_comparison}.

The results show that our method consistently achieves the best performance in terms of FID, $\mathrm{FID}_{\mathrm{CLIP}}$, and CLIP-I on both the pose-agnostic and pose-different subsets. For most baselines, performance degrades under this setting, likely because they are neither trained nor designed to operate on hair-removed source images, and thus face a distribution shift when the original source image is replaced with a bald base. At the same time, the FLUX-based hair-removed input does appear to provide a favorable condition for preserving non-hair regions, as reflected by the substantial gains of Stable Hair~\cite{zhang2025stable} in SSIM and PSNR, together with noticeable improvements in FID and $\mathrm{FID}_{\mathrm{CLIP}}$. However, these gains do not translate into corresponding improvements in reference-hairstyle transfer quality or overall perceptual quality, where our method remains superior. Considering that our method is designed not primarily to preserve the source non-hair appearance itself, but rather to generate a hairstyle aligned with the source head geometry and pose, these results suggest that the advantage of our method cannot be explained solely by the use of FLUX-based hair-removed inputs.

\begin{figure}[t]
    \centering
    \includegraphics[width=\linewidth]{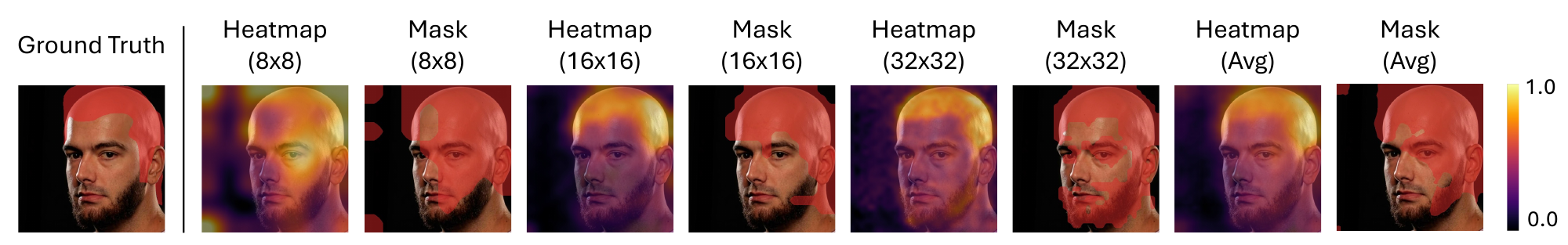}
    \caption{Analysis of attention resolution for mask extraction. The leftmost image shows the source image overlaid with the ground-truth hair mask. The remaining columns show, from left to right, the attention heat map and the resulting mask overlaid on the source image, extracted from the $8\times8$, $16\times16$, $32\times32$, and all-resolution averaged cross-attention maps, respectively. The $16\times16$ attention maps provide an appropriate granularity for coarse hair localization.}
    \label{sup:fig:why_16_res_attn}
\end{figure}

\subsection{Effect of Training Data Composition}
\label{sup:quant:dataset}

We further analyze the effect of including CelebV-HQ~\cite{zhu2022celebv} in training.
In our training setup, FFHQ~\cite{karras2019style} samples are formed as self-pairs where the same image serves as both the reference and the target ($x_j=x_i$), whereas CelebV-HQ provides cross-frame pairs where $x_j$ and $x_i$ correspond to different frames of the same identity ($x_j\neq x_i$).
Our full model is trained on both FFHQ and CelebV-HQ, whereas the variant \textit{Ours w/o CelebV-HQ} is trained on FFHQ only. Results on the pose-agnostic and pose-different subsets are presented in \cref{tab:train_data_comp}.

Overall, adding CelebV-HQ improves SSIM and PSNR on both subsets, indicating better preservation of source structure.
We attribute this gain to the cross-frame supervision in CelebV-HQ: since reference and target frames differ in pose and expression while sharing the same identity, the model is encouraged to learn stronger invariances in non-hair regions and to avoid unnecessary changes beyond the intended hair manipulation, leading to improved structural fidelity.

For the fidelity metrics, \textit{Ours w/o CelebV-HQ} yields lower FID, whereas the full model yields lower FID$_{\mathrm{CLIP}}$, indicating a metric-dependent trade-off under different training data compositions.
Notably, CLIP-I remains comparable across the two training-data compositions on both subsets. This suggests that overall fidelity exhibits a metric-dependent trade-off (FID vs.\ FID$_{\mathrm{CLIP}}$) with the inclusion of cross-frame pairs, while self-paired supervision alone can still capture reference hairstyle attributes to a large extent.

\begin{figure}[t]
    \centering
    \includegraphics[width=0.6\linewidth]{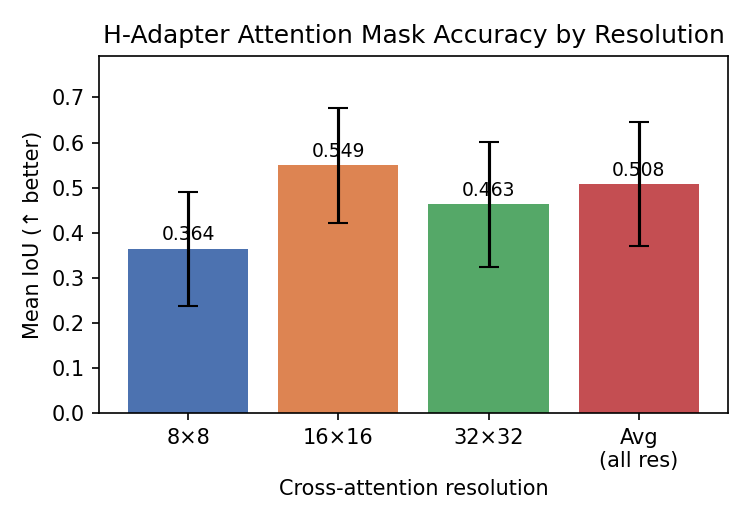}
    \caption{Quantitative comparison of attention resolutions for mask extraction. Evaluated on 3{,}000 randomly sampled images, the bars show the mean IoU between the attention-derived mask and the ground-truth hair mask, with error bars representing the standard deviation across samples. The $16\times16$ resolution achieves the highest mean IoU, supporting its use for coarse mask extraction.}
    \label{sup:fig:iou_barchart}
\end{figure}

\section{Analysis of Attention-Based Mask Extraction}
\label{sup:attn}

\subsection{Choice of Attention Resolution for Mask Extraction}
\label{sup:attn:why_16_res_attn}

As described in the main paper, we derive an attention-based coarse mask by aggregating cross-attention maps over all tokens except the separator token, which consistently attends to non-hair regions. This mask is used as a coarse spatial prior for inpainting and attention gating.

This subsection examines which attention resolution is most suitable for mask extraction. We compare masks extracted from $8\times8$, $16\times16$, $32\times32$, and all resolutions combined. For controlled evaluation, we use the same image as the reference, while the source is constructed as a hair-removed base image of the same sample. This setting provides a well-defined ground-truth hair mask for IoU computation; with different source and reference images, the target hair region would not be uniquely defined. We then extract the mask after a one-step warm-up inference. For each resolution, we aggregate the cross-attention maps and obtain a binary mask by thresholding the resulting attention map at its mean value. To isolate the effect of attention resolution itself, we do not apply any additional post-processing to the resulting mask.

As shown in \cref{sup:fig:why_16_res_attn}, the $16\times16$ attention maps provide an appropriate granularity for coarse hair localization, capturing the overall spatial extent of hair without being excessively coarse or dominated by fine high-frequency structures. This observation is further supported by the quantitative comparison in \cref{sup:fig:iou_barchart}, where the $16\times16$ resolution achieves the highest mean IoU with the ground-truth hair mask over 3{,}000 randomly sampled images. Based on these qualitative and quantitative results, we use the $16\times16$ cross-attention maps for mask extraction throughout the paper.

\begin{figure}[t]
    \centering
    \includegraphics[width=\linewidth]{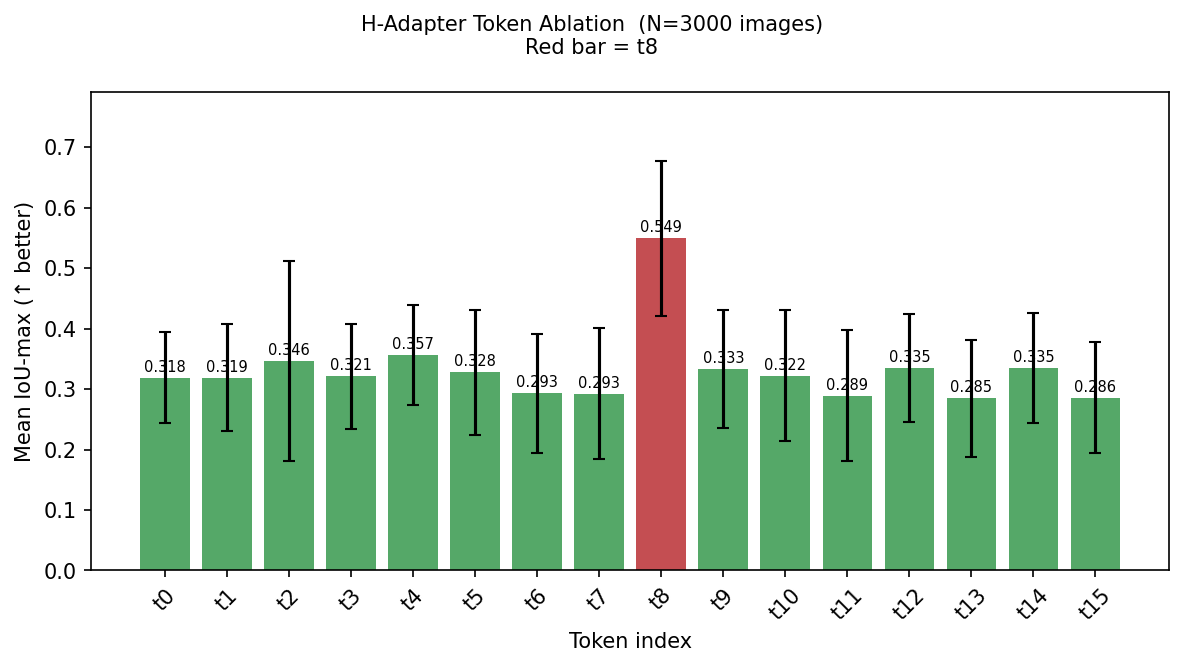}
    \caption{Separator-token analysis over the 16 IP-Adapter tokens. For each token, we retain the larger IoU between the mask obtained by excluding that token from attention aggregation and the corresponding token-only mask. The highest mean IoU is achieved by $t_8$ over 3{,}000 samples, supporting its use as the separator token in our pipeline.}
    \label{sup:fig:separator_token_iou}
\end{figure}

\subsection{Choice of the Separator Token}
\label{sup:attn:separator_token}

After fixing the attention resolution for mask extraction, we next examine which IP-Adapter token most consistently separates hair from non-hair regions.

Our pipeline constructs the coarse attention mask as
$M_{\mathrm{attn}}=\mathrm{Binarize}\!\left(\sum_{k\neq s} CA(t_k)\right)$, where $CA(t_k)$ denotes the cross-attention map of token $t_k$, and $s$ denotes the index of the separator token.

To determine the separator token, we compare all 16 tokens as candidates under the same controlled evaluation protocol as in \cref{sup:attn:why_16_res_attn}. For each candidate separator index $s\in\{0,\dots,15\}$, we construct the coarse mask using the definition above and also evaluate the token-only mask $\mathrm{Binarize}\!\left(CA(t_s)\right)$ against the ground-truth hair mask. We then report the larger IoU between the two masks, which measures how well each token separates hair and non-hair regions without assuming in advance whether the token itself corresponds to hair or to non-hair.

As shown in \cref{sup:fig:separator_token_iou}, $t_8$ achieves the highest mean $\mathrm{IoU}_{\max}$ ($0.549$) over 3{,}000 randomly sampled images, while the remaining tokens yield similar but consistently lower values. Based on this result, we use $s=8$ as the separator token in the main pipeline.

\begin{figure}[t]
\centering
\includegraphics[width=0.8\linewidth]{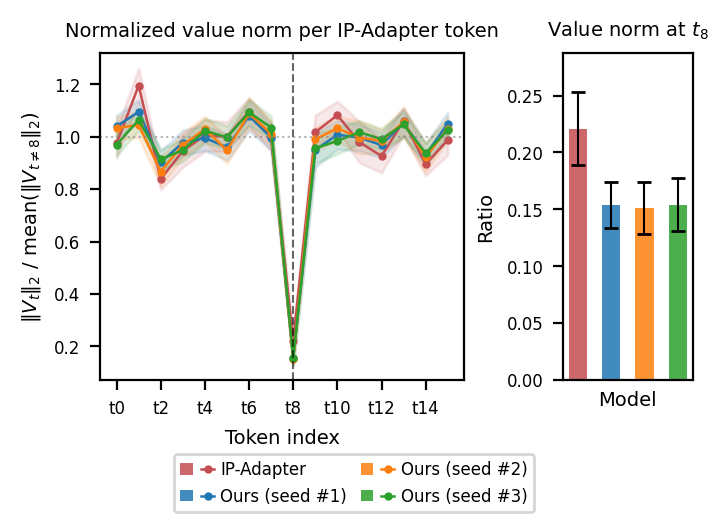}
\caption{Analysis of separator-token behavior. We visualize the normalized value-vector norms of IP-Adapter and H-Adapter tokens. The $t_8$ token already has a relatively small norm in the pretrained IP-Adapter, and the proposed region-specific objective further reduces its relative norm in H-Adapter. This supports the interpretation that $t_8$ acts as a stable non-hair separator rather than a single-run artifact.}
\label{fig:sup:separator_norm}
\end{figure}

\subsection{Value-Vector Analysis of the Separator Token}
We further analyze the separator-token behavior from the perspective of the adapter value vectors.
For each token, we compute its normalized value-vector norm by dividing its norm by the average norm of the remaining tokens.
As shown in \cref{fig:sup:separator_norm}, the pretrained IP-Adapter already assigns the smallest value-vector norm to $t_8$, suggesting that this token has a relatively weak role in injecting reference appearance.
After training with the proposed region-specific objective, H-Adapter further reduces the relative norm of $t_8$ from approximately $0.22$ to $0.15$ of the other-token average, while its attention map consistently aligns with non-hair regions across different random seeds.
This suggests that the region-specific objective amplifies a weak pre-existing token-level bias into a stable non-hair separator, rather than producing a single-run artifact.

\subsection{Comparison of Attention-Derived Masks with HairFusion}
\label{sup:attn:compar}

We further compare masks derived from cross-attention in our method and HairFusion~\cite{chung2025preserve} to analyze how the training design affects the extracted masks and the resulting generation quality. To compare the cross-attention-derived masks directly, we visualize only the mask extracted from cross-attention for each method. In HairFusion, this mask is not the final mask used for latent blending, as it is further combined with the source hair mask in the original pipeline. The visualized masks also arise at different stages of the two pipelines, reflecting the design of each method: for our method, the mask is extracted from the single warm-up denoising step, whereas in HairFusion, latent blending is performed during the last $n$ denoising steps, with a cross-attention-based mask extracted at each step; here we show the mask from the final step $T$.

\begin{figure}[t]
    \centering
    \includegraphics[width=\linewidth]{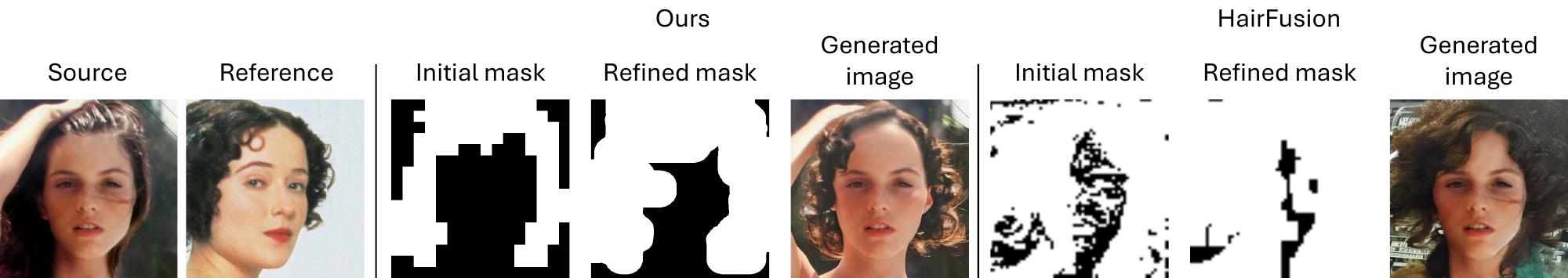}
    \caption{Comparison of attention-derived masks and generated results between our method and HairFusion~\cite{chung2025preserve}. The initial mask is obtained by thresholding the attention-derived map, while the refined mask denotes the post-processed mask actually used in each pipeline. Our method yields an initial mask that more selectively captures the hair region, making it better suited for spatially localizing hairstyle transfer.}
    \label{sup:fig:attn_compare}
\end{figure}

As shown in \cref{sup:fig:attn_compare}, although both methods use cross-attention to obtain spatial guidance, the resulting masks differ in their selectivity and localization. Both HairFusion and our method demonstrate that cross-attention can provide useful spatial cues for hairstyle transfer. Our method further improves the selectivity of this signal through region-specific training, encouraging H-Adapter to transfer reference hairstyle information primarily within hair regions while limiting its effect on non-hair regions. As a result, the coarse mask derived from cross-attention is more tightly localized to hair regions, whereas the mask from HairFusion tends to extend more broadly into nearby visual context.

\section{Discussion and Future Work}
\label{sup:discussion}

\subsubsection{Extension to newer backbones.}
H-Adapter is implemented and validated on the SD1.5/IP-Adapter-Plus backbone. Generalizing the framework to newer diffusion backbones and adapter architectures is an important direction for future work. Since token behavior and attention patterns can differ across backbone and adapter designs, this motivates future work on adapting the separator-token selection and attention-based mask extraction strategy to each target architecture.

\subsubsection{Out-of-Benchmark Occlusions.}
Our benchmark focuses on hairstyle transfer for face images without explicit head coverings. Cases involving hats, scarves, helmets, or other head coverings are therefore outside the scope of the current benchmark. While our additional qualitative results show that H-Adapter can handle diverse and unusual hairstyles, robustly distinguishing hair from non-hair head coverings remains an open challenge. Extending the benchmark and method to such occluded cases is a promising direction for future work.

\clearpage
\section{FLUX Editing Prompt for Hair-Removed Base Image Generation}
\label{sup:sec:flux_prompt}

This section provides the exact prompt templates used in our experiments.
{\scriptsize
\begin{Verbatim}[
    breaklines=true,
    breakanywhere=true,
    breaksymbolleft={},
    breaksymbolright={}
]
Using the provided reference image, remove all hair so the person is completely bald. All non-hair regions are preserved exactly as in the original image: identical pixels, identical color channels, unchanged lighting, background, facial identity, pose, composition, and focal characteristics, with no global or local color shift. Edits are confined strictly to the original hair region only. The result contains no new elements: no added objects, accessories, extra details, or stylization. Image rendering remains unchanged everywhere else.
\end{Verbatim}
}
\section{Source and License Information}
\label{sup:source-license}

\begin{table}[t]
\centering
\caption{Source and license information for third-party stock images and illustrations used in qualitative examples.}
\label{tab:sup:stock-sources}
\resizebox{\linewidth}{!}{
\begin{tabular}{lllll}
\toprule
Figure & Source & ID & Creator & License \\
\midrule
\cref{fig:in-the-wild} & Pixabay &
8982528 & Frank\_Rietsch & Pixabay Content License \\
\cref{fig:in-the-wild} & Unsplash &
kxLQAIL724k & Luis Olmos & Unsplash License \\
\cref{fig:in-the-wild} & Unsplash &
16rhnOIB3TY & Daria Strategy & Unsplash License \\
\cref{sup:fig:diverse_domain} & Pixabay &
8379544 & esvisionaria & Pixabay Content License \\
\cref{sup:fig:diverse_domain} & Pixabay &
8111624 & aahihi & Pixabay Content License \\
\cref{sup:fig:diverse_domain} & Pixabay &
7967243 & bodlo & Pixabay Content License \\
\cref{sup:fig:diverse_domain} & Pixabay &
7951986 & bodlo & Pixabay Content License \\
\cref{fig:in-the-wild} &
Pexels & Various & Various photographers & Pexels License \\
\cref{sup:fig:diverse_domain,sup:fig:results:in_the_wild:failure,sup:fig:results:t2i,sup:fig:results:plus} &
Pexels & Various & Various photographers & Pexels License \\
\bottomrule
\end{tabular}
}
\end{table}

\Cref{tab:sup:stock-sources} lists the source and license information
for third-party stock images and illustrations used in the qualitative examples.
All listed images were used only as source/reference inputs for qualitative
hairstyle-transfer visualization.

\end{document}